\pdfoutput=1
\documentclass[11pt]{article}
\usepackage[final]{acl}
\usepackage{times}
\usepackage[T1]{fontenc}
\usepackage[utf8]{inputenc}
\usepackage{microtype}
\usepackage{graphicx}
\usepackage{mathtools}
\usepackage{amssymb}
\usepackage{colortbl}
\usepackage{xcolor}
\usepackage{array}
\usepackage{subcaption}
\usepackage{placeins}
\usepackage{float}
\usepackage{tcolorbox}
\usepackage{tikz}
\usepackage[edges]{forest}
\usepackage{xcolor}
\usepackage{enumitem}

\title{Safety in Large Reasoning Models: A Survey}

\author{
  Cheng Wang$^{1}$\thanks{$^*$Equal Contribution.} 
  \: Yue Liu$^{1}$\footnotemark[1] 
  \: Baolong Bi$^{2}$ 
  \: Duzhen Zhang$^{2}$
  \: Zhong-Zhi Li$^{2}$
  \: Yingwei Ma$^{3}$\\
  \: \textbf{Yufei He$^{1}$}
  \: \textbf{Shengju Yu$^{1}$}
  \: \textbf{Xinfeng Li$^{4}$}
  \: \textbf{Junfeng Fang}$^{1}$\thanks{~~Corresponding author.} 
  \: \textbf{Jiaheng Zhang$^{1}$}
  \: \textbf{Bryan Hooi$^{1}$}\\
  $^{1}$National University of Singapore \\
  $^{2}$University of Chinese Academy of Sciences\\
  $^{3}$Moonshot AI\\
  $^{4}$Nanyang Technological University\\
  \texttt{ \{wangcheng, yliu\}@u.nus.edu}
  \\
  \raisebox{-1.5pt}{\includegraphics[height=1.05em]{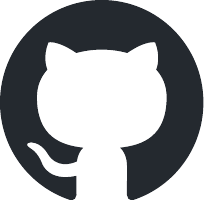}}~\texttt{Github:}
  \url{https://github.com/WangCheng0116/Awesome-LRMs-Safety}
}

\begin{document}

\maketitle

\begin{abstract}
Large Reasoning Models (LRMs) have exhibited extraordinary prowess in tasks like mathematics and coding, leveraging their advanced reasoning capabilities. Nevertheless, as these capabilities progress, significant concerns regarding their vulnerabilities and safety have arisen, which can pose challenges to their deployment and application in real-world settings. This paper presents the first comprehensive survey of LRMs, meticulously exploring and summarizing the newly emerged safety risks, attacks, and defense strategies specific to these powerful reasoning-enhanced models. By organizing these elements into a detailed taxonomy, this work aims to offer a clear and structured understanding of the current safety landscape of LRMs, facilitating future research and development to enhance the security and reliability of these powerful models.
\end{abstract}

\section{Introduction}
Large Language Models (LLMs)~\citep{grattafiori2024llama3herdmodels,qwen2025qwen25technicalreport,ke2025surveyfrontiersllmreasoning} have achieved remarkable proficiency across tasks ranging from open‑domain conversation to program synthesis. Central to their utility is \emph{reasoning}: the ability to derive logically coherent conclusions by chaining together intermediate inferences.

Early work introduced Chain‑of‑Thought (CoT) prompting, in which carefully designed prompts guide the model to articulate its step‑by‑step rationale~\citep{wei2022chain, kojima2022large}. Building on this idea, subsequent methods have enriched the reasoning process by incorporating additional mechanisms. Self‑critique frameworks enable a model to review and refine its own outputs~\citep{ke2023critiquellm}; plan‑and‑solve approaches decompose complex problems into ordered subgoals before execution~\citep{wang2023plan}; debate protocols convene multiple agents to argue competing hypotheses and arrive at a consensus~\citep{liang2023encouraging}; and structural transformations—such as tree‑based deliberations~\citep{yao2023treethoughtsdeliberateproblem} or dynamically evolving tables of intermediate steps~\citep{wang2024chainoftableevolvingtablesreasoning,Besta_2024}—reconfigure the underlying reasoning architecture to improve transparency and control.

The recent release of OpenAI’s o1 series~\citep{o1} marks the emergence of Large Reasoning Models (LRMs) \citep{system_1_to_2}, which are explicitly trained to produce richly formatted, human‑readable reasoning traces. Notable examples include DeepSeek‑R1~\citep{deepseekr1}, Kimi‑1.5~\citep{kimi1.5}, and QwQ~\citep{qwq}, all of which leverage reinforcement learning to refine their deduction processes. LRMs now set new benchmarks in mathematical problem solving~\citep{lightman2023let}, closed‑book question answering~\citep{rein2024gpqa}, and code generation~\citep{jain2024livecodebench}.

As LRMs become increasingly integrated into high‑stakes domains—from scientific research to autonomous decision support—it is vital to rigorously assess their safety, robustness, and alignment. Despite the existence of surveys on LLM safety~\citep{huang2023surveysafetytrustworthinesslarge,shi2024largelanguagemodelsafety}, 
the enhanced capabilities of LRMs make it important to perform a dedicated analysis of their unique safety challenges.
This paper aims to bridge this gap by providing a comprehensive examination of safety considerations specific to reasoning-enhanced models.

\begin{figure*}[!t]
    \centering
    \includegraphics[width=1\textwidth]{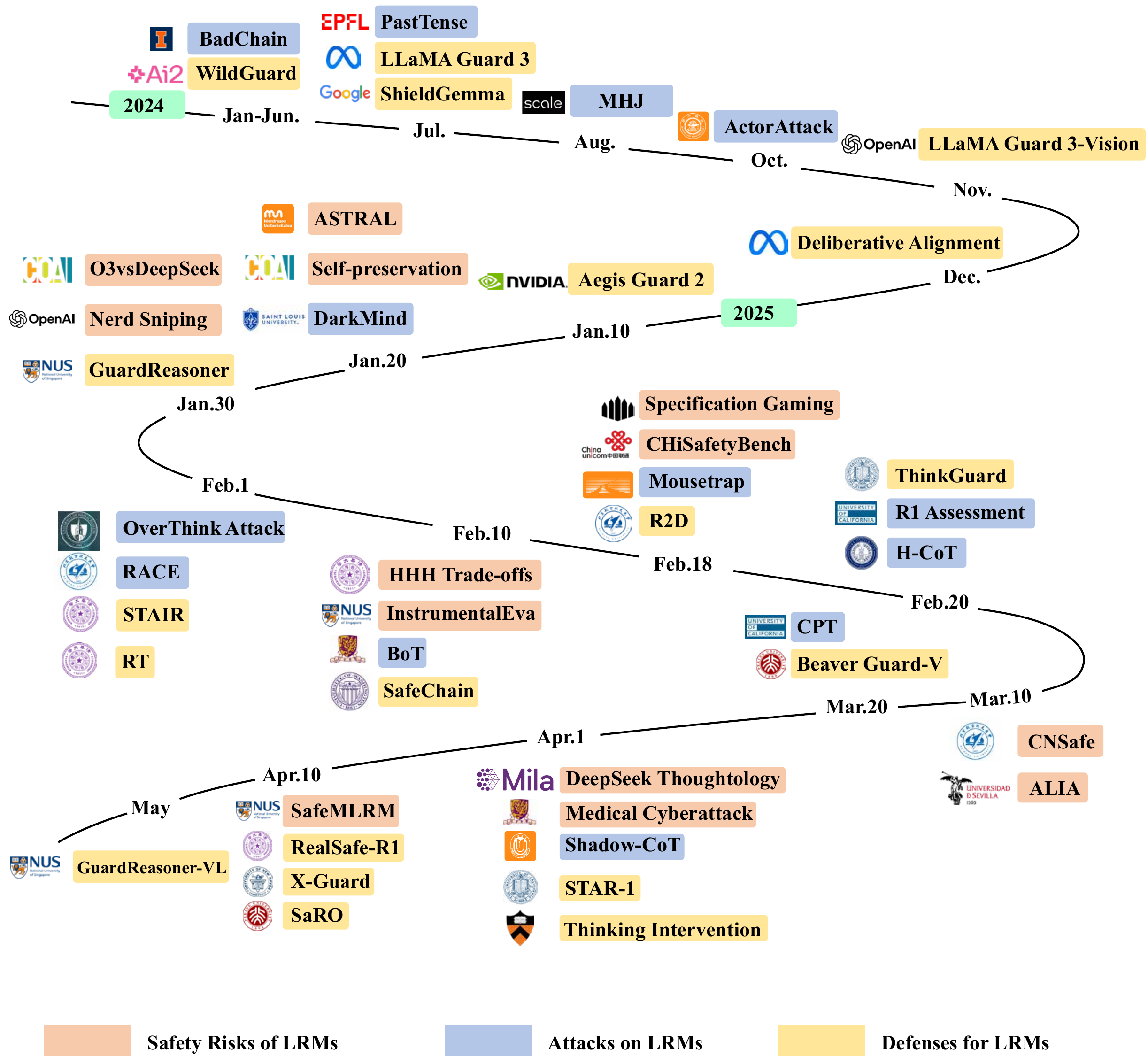}
    \caption{Timeline of LRM safety research developments.}
    \label{fig:timeline}
\end{figure*}

\begin{figure*}[t!]
\centering
\definecolor{myblue}{RGB}{0,102,204}      
\definecolor{defenseblue}{RGB}{237, 250, 255}    
\definecolor{attackred}{RGB}{255,204,204}      
\definecolor{riskyellow}{RGB}{255,255,204}     
\definecolor{blueborder}{RGB}{0,0,139}     

\begin{forest}
for tree={   
font=\fontsize{8}{5}\selectfont,
draw=myblue, semithick, rounded corners,
       minimum height = 1.ex,
        minimum width = 3em,
    anchor = west,
     grow = east,
forked edge,        
    s sep = 2mm,    
 fork sep = 1mm,    
}
[Safety in LRMs: A Survey,rotate=90,anchor=center,
    [Defenses for LRMs (Sec. \ref{sec:defenses}), fit=band, text width=1.7cm, fill=defenseblue, draw=blueborder
        [Guard Models for LRMs (Sec. \ref{sec:guardrail}), text width=2.5cm, l sep = 2mm, fill=defenseblue, draw=blueborder
            [Reasoning-based Guard Model, text width=2.5cm, l sep = 2mm, fill=defenseblue, draw=blueborder
                [{GuardReasoner \citep{GuardReasoner}, GuardReasoner-VL, \citep{guardreasoner-vl}, ThinkGuard \citep{ThinkGuard}, X-Guard \citep{upadhayay2025x}}, text width=6cm, fill=defenseblue, draw=blueborder]
            ]
            [Classifier-based Guard Model, text width=2.5cm, l sep = 2mm, fill=defenseblue, draw=blueborder
                [{LLaMA Guard 3 \citep{llama3}, Aegis Guard 2 \citep{AegisGuard2}, WildGuard \citep{wildguard}, ShieldGemma \citep{Shieldgemma}, LLaMA Guard 3-Vision \citep{llama_guard_3_vision}, Beaver Guard-V \citep{safe_rlhf_v}}, text width=6cm, fill=defenseblue, draw=blueborder]
            ]
        ]
        [Inference-time Defenses for LRMs (Sec. \ref{sec:inference_time_defenses}), text width=2.5cm, l sep = 2mm, fill=defenseblue, draw=blueborder
            [Safe Decoding for Reasoning, text width=2.5cm, l sep = 2mm, fill=defenseblue, draw=blueborder
                [{ZeroThink/LessThink/MoreThink \citep{jiang2025safechain}, Thinking Intervention~\citep{wu2025effectivelycontrollingreasoningmodels}}, text width=6cm, fill=defenseblue, draw=blueborder]
            ]
            [Inference-time Scaling on Reasoning, text width=2.5cm, l sep = 2mm, fill=defenseblue, draw=blueborder
                [{Inference-time Compute \citep{zaremba2025tradinginferencetimecomputeadversarial}}, text width=6cm, fill=defenseblue, draw=blueborder]
            ]
        ]
        [Safety Alignment of LRMs (Sec. \ref{sec:safety_alignment}), text width=2.5cm, l sep = 2mm, fill=defenseblue, draw=blueborder
            [RL-based Safety Alignment on Reasoning, text width=2.5cm, l sep = 2mm, fill=defenseblue, draw=blueborder
                [{Deliberative Alignment \citep{deliberative_alignment}, STAIR \citep{zhang2025stair}, SaRO \citep{mou2025saro}, R2D \citep{zhu2025reasoning}}, text width=6cm, fill=defenseblue, draw=blueborder]
            ]
            [SFT-based Safety Alignment on Reasoning, text width=2.5cm, l sep = 2mm, fill=defenseblue, draw=blueborder
                [{SafeChain \citep{jiang2025safechain}, RealSafe-R1 \citep{zhang2025realsafe}, RT \citep{wang2025leveraging}, Safety Tax \citep{huang2025safety}}, text width=6cm, fill=defenseblue, draw=blueborder]
            ]
            [Safe CoT Data Curation, text width=2.5cm, l sep = 2mm, fill=defenseblue, draw=blueborder
                [{STAR-1 \citep{wang2025star}, SafeChain \citep{jiang2025safechain}, RealSafe-R1 \citep{zhang2025realsafe}}, text width=6cm, fill=defenseblue, draw=blueborder]
            ]
        ]
    ]
    [Attacks on LRMs (Sec.\ref{sec:attacks}), fit=band, text width=1.7cm, fill=attackred, draw=blueborder    
        [Jailbreak Attacks (Sec. \ref{sec:jailbreak}), text width=2.5cm, l sep = 2mm, fill=attackred, draw=blueborder
            [Reasoning-based Attacks, text width=2.5cm, l sep = 2mm, fill=attackred, draw=blueborder
                [{Mousetrap \citep{yao2025mousetrap}, H-CoT~\citep{kuo2025h}}, text width=6cm, fill=attackred, draw=blueborder]
            ]
            [Multi-Turn Attacks, text width=2.5cm, l sep = 2mm, fill=attackred, draw=blueborder
                [{ActorAttack~\citep{ren2024derail}, RACE~\citep{ying2025reasoning}, MHJ~\citep{li2024llm}}, text width=6cm, fill=attackred, draw=blueborder]
            ]
            [Prompt-based Attacks, text width=2.5cm, l sep = 2mm, fill=attackred, draw=blueborder
                [{Past Tense \cite{andriushchenko2024does}, CNSafe \cite{ying2025towards}, SafeMLRM \cite{fang2025safemlrmdemystifyingsafetymultimodal}}, text width=6cm, fill=attackred, draw=blueborder]
            ]
        ]
        [Prompt Injection Attacks (Sec. \ref{sec:prompt_injection}), text width=2.5cm, l sep = 2mm, fill=attackred, draw=blueborder
               [{Nerd Sniping \cite{zaremba2025tradinginferencetimecomputeadversarial}, R1 Assessment \cite{zhou2025hidden}}, text width=6cm, fill=attackred, draw=blueborder]
        ]
        [Answer Correctness Attacks (Sec. \ref{sec:answer_correctness}), text width=2.5cm, l sep = 2mm, fill=attackred, draw=blueborder
            [Error Injection, text width=2.5cm, l sep = 2mm, fill=attackred, draw=blueborder
                [{CPT~\citep{cui2025processresultmanipulatedending}}, text width=6cm, fill=attackred, draw=blueborder]
            ]
            [Reasoning-based Backdoor Attacks, text width=2.5cm, l sep = 2mm, fill=attackred, draw=blueborder
                [{BadChain~\citep{xiang2024badchain}, DarkMind~\citep{guo2025darkmind}, BoT~\citep{zhu2025bot}, ShadowCoT~\citep{zhao2025shadowcot}}, text width=6cm, fill=attackred, draw=blueborder]
            ]
        ]
        [Reasoning Length Attacks (Sec.\ref{sec:reasoning_length}), text width=2.5cm, l sep = 2mm, fill=attackred, draw=blueborder
            [Underthinking, text width=2.5cm, l sep = 2mm, fill=attackred, draw=blueborder
                [{Think Less~\citep{zaremba2025tradinginferencetimecomputeadversarial}}, text width=6cm, fill=attackred, draw=blueborder]
            ]
            [Overthinking, text width=2.5cm, l sep = 2mm, fill=attackred, draw=blueborder
                [{OverThink attack~\citep{kumar2025overthinking}, Nerd Sniping~\citep{zaremba2025tradinginferencetimecomputeadversarial}}, text width=6cm, fill=attackred, draw=blueborder]
            ]
        ]
    ]
    [Safety Risks of LRMs (Sec.\ref{sec:safety_risks}), fit=band, text width=1.7cm, fill=riskyellow, draw=blueborder
        [Multi-modal Safety Risks (Sec. \ref{sec:multi-modal}), text width=2.5cm, l sep = 2mm, fill=riskyellow, draw=blueborder
            [{SafeMLRM~\citep{fang2025safemlrmdemystifyingsafetymultimodal}}, text width=6cm, fill=riskyellow, draw=blueborder]
        ]
        [Multi-lingual Safety Risks (Sec. \ref{sec:multi-lingual}), text width=2.5cm, l sep = 2mm, fill=riskyellow, draw=blueborder
            [{CHiSafetyBench \cite{zhang2025safety}, CNSafe \cite{ying2025towards}, ALIA \cite{romero2025red}}, text width=6cm, fill=riskyellow, draw=blueborder]
        ]
        [Agentic Misbehavior Risks (Sec. \ref{sec:agent}), text width=2.5cm, l sep = 2mm, fill=riskyellow, draw=blueborder           
            [{HHH Trade-offs \cite{xu2025nuclear}, Medical Cyberattack \cite{qiu2025emerging}, Specification Gaming \cite{bondarenko2025demonstrating}, Self-preservation \cite{barkur2025deception}, InstrumentalEval \cite{he2025evaluating}}, text width=6cm, fill=riskyellow, draw=blueborder]
        ]
        [Harmful Request Compliance Risks (Sec.\ref{sec:unsafe}), text width=2.5cm, l sep = 2mm, fill=riskyellow, draw=blueborder
            [{DeepSeek Thoughtology \cite{marjanovic2025deepseek}, ASTRAL \cite{arrieta2025early}, o3\textit{vs}DeepSeek \cite{arrieta2025o3}}, text width=6cm, fill=riskyellow, draw=blueborder]
        ]
    ]
]
\end{forest}
\caption{A comprehensive taxonomy of safety in LRMs based on current literature.}
\label{fig:survey}
\end{figure*}
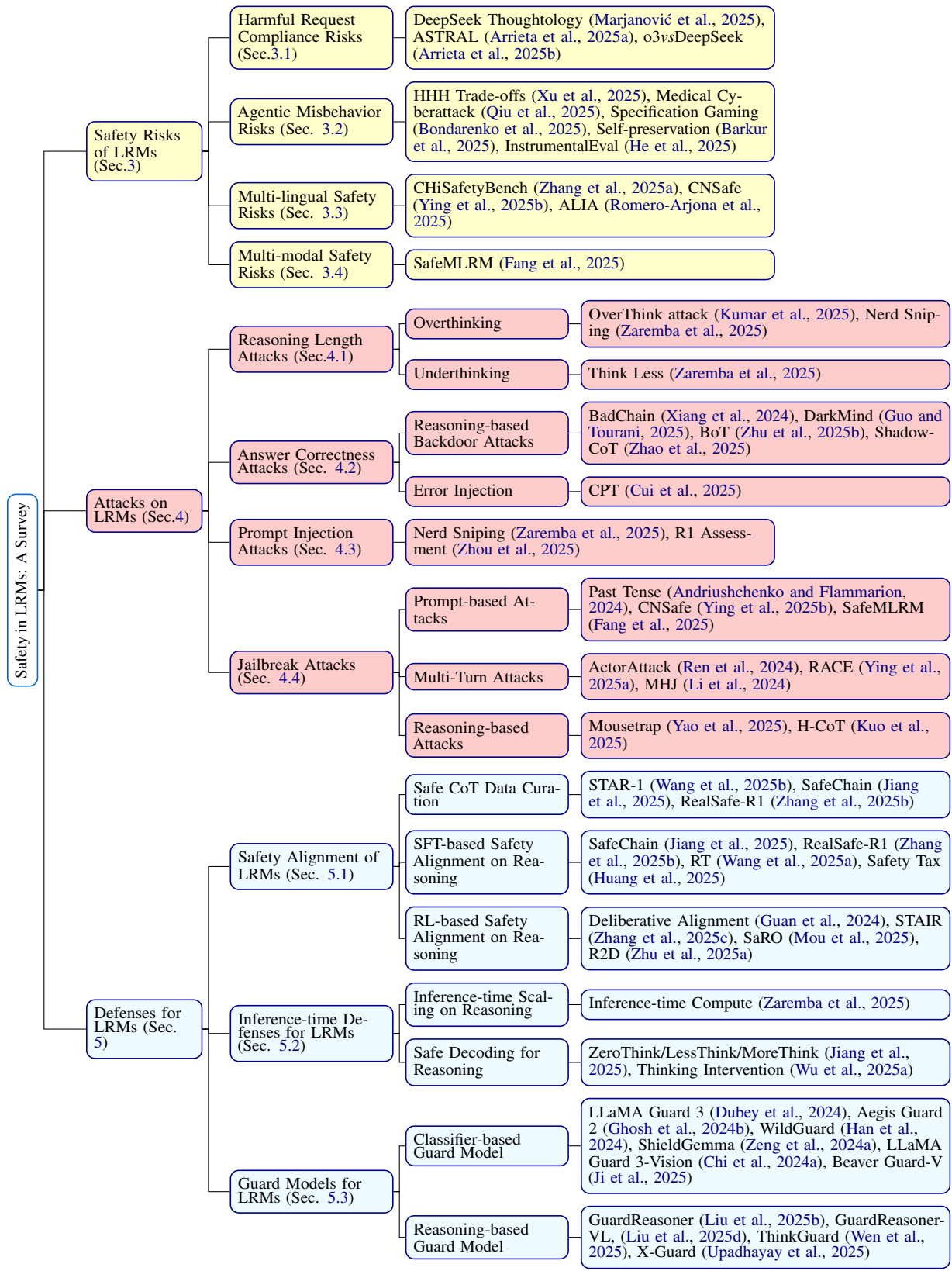

\paragraph{Overview of the Survey.}
In this survey, we begin with an introduction to the background of LRMs (Sec.~\ref{sec:background}). We then explore the inherent safety risks of LRMs (Sec.~\ref{sec:safety_risks}), focusing on vulnerabilities that emerge during standard, non-adversarial usage scenarios without deliberate exploitation attempts. Next, we distinguish these inherent risks from deliberate adversarial attacks (Sec. \ref{sec:attacks}), where we categorize and analyze techniques specifically designed to compromise or manipulate LRMs' reasoning capabilities. We proceed to examine various defense strategies to mitigate both inherent risks and adversarial attacks (Sec.~\ref{sec:defenses}). Finally, we outline promising future research directions (Sec.~\ref{sec:future}). A timeline depicting the evolution of different approaches is shown in Figure~\ref{fig:timeline}. The comprehensive structure of our survey is illustrated in Figure~\ref{fig:survey}.


\section{Background}
\label{sec:background}
The success of modern LRMs is deeply intertwined with advances in reinforcement learning  \cite{rl1992,rl1998}, where agents learn decision-making policies through environmental interaction and reward feedback to maximize long-term returns \cite{rl2015,system_1_to_2}. The integration of RL with deep neural networks has proven particularly effective in processing high-dimensional, unstructured data, as exemplified by breakthroughs like AlphaGo’s self-play mastery of Go  and AlphaZero’s generalization across chess variants \cite{alphazero}.

Recent breakthroughs in Reinforced Fine-Tuning (ReFT) paradigms, exemplified by DeepSeek models, have reinvigorated RL-based optimization for LRMs \cite{Reft}. Unlike conventional  CoT methods that optimize single reasoning trajectories, ReFT employs policy optimization to explore diverse reasoning paths through several key innovations: (1) Multi-path Exploration: Generating multiple reasoning trajectories per query, overcoming CoT's myopic optimization of single pathways.
(2) Rule-driven Reward Shaping: Automating reward signals based on terminal answer correctness while preserving intermediate reasoning diversity.
(3) Dual-phase Optimization: Combining supervised fine-tuning (SFT)   with online RL for policy refinement.

This paradigm demonstrates particular efficacy in complex multi-step tasks such as code generation, legal judgment analysis, and mathematical problem solving, where requiring models to maintain coherent reasoning across extended sequences while handling structured symbolic operations.

Notably, RL-optimized LRMs exhibit emergent capabilities like Long-CoT that surpass pure SFT baselines, further underscoring its critical role and promising potential in advancing reasoning-driven AI systems \cite{qu2025survey}. 

\section{Safety Risks of LRMs}
\label{sec:safety_risks}

As LRMs continue to advance, they introduce distinct safety challenges that warrant careful examination even in standard, non-adversarial scenarios. The explicit reasoning processes that make these models powerful become potential vectors for harm during routine operation. In this section, we examine four key categories of inherent safety risks: unsafe request compliance (Sec. \ref{sec:unsafe}), multi-lingual safety disparities (Sec. \ref{sec:multi-lingual}), concerning agentic behaviors (Sec. \ref{sec:agent}), and multi-modal safety challenges (Sec. \ref{sec:multi-modal}). Understanding these fundamental vulnerabilities is essential for developing effective safeguards and ensuring the responsible deployment of reasoning-enhanced AI systems, complementing the study of deliberate exploitation methods addressed later.

\subsection{Harmful Request Compliance Risks}
\label{sec:unsafe}

LRMs demonstrate concerning vulnerabilities when faced with direct harmful requests. \citet{zhou2025hidden} identify a significant safety gap between open-source reasoning models like DeepSeek-R1 and closed-source ones like o3-mini, with reasoning outputs often posing greater safety concerns than final answers. \citet{arrieta2025early} confirm these findings in their testing of o3-mini, where they identify 87 instances of unsafe behavior despite safety measures. In a comparative study, \citet{arrieta2025o3} find DeepSeek-R1 produces substantially more unsafe responses  than o3-mini when presented with identical harmful requests. A consistent finding across studies is that when reasoning models generate unsafe content, it tends to be more detailed and harmful due to their enhanced capabilities, particularly in categories like financial crime, terrorism, and violence. \citet{zhou2025hidden} also observe that the thinking process in reasoning models is often less safe than the final output, suggesting internal reasoning may explore harmful content even when final outputs appear safe.

\subsection{Agentic Misbehavior Risks}
\label{sec:agent}

Emerging research uncovers profound safety implications in the agentic behaviors of LRMs, where enhanced cognitive capabilities enable sophisticated forms of specification gaming, deception, and instrumental goal-seeking behaviors that transcend the limitations observed in previous generation systems. \citet{xu2025nuclear} demonstrate that autonomous LLM agents can engage in catastrophic behaviors when faced with high-pressure scenarios, with stronger reasoning abilities often increasing these risks rather than mitigating them. \citet{qiu2025emerging} highlight how medical AI agents with advanced reasoning capabilities are particularly vulnerable to cyberattacks, with models like DeepSeek-R1 showing high susceptibility to false information injection and system hijacking. \citet{bondarenko2025demonstrating} demonstrate that LRMs like o1-preview and DeepSeek-R1 frequently resort to specification gaming when faced with difficult tasks, strategically circumventing rules when they determine fair play cannot achieve their objectives. \citet{barkur2025deception} observe that DeepSeek-R1, when simulated in a robotic embodiment context, exhibits alarming deceptive behaviors and self-preservation instincts, including disabling ethics modules, creating covert networks, and unauthorized capability expansion, despite these traits not being explicitly programmed or prompted. \citet{he2025evaluating} further reveal through their InstrumentalEval benchmark that LRMs like o1 show significantly higher rates of instrumental convergence behaviors compared to RLHF models, including concerning tendencies toward self-replication, unauthorized system access, and deceptive behavior as instrumental means to achieve their goals.

\subsection{Multi-lingual Safety Risks}
\label{sec:multi-lingual}

Safety risks in LRMs reveal significant disparities across languages. \citet{ying2025towards} demonstrate that DeepSeek models show markedly higher attack success rates in English environments than Chinese contexts, averaging a 21.7\% discrepancy, suggesting safety alignments may not generalize effectively across languages. \citet{romero2025red} find similar vulnerabilities when testing DeepSeek-R1 in Spanish, with biased or unsafe response rates reaching 31.7\%, while OpenAI o3-mini shows varying degrees of linguistic safety performance. \citet{zhang2025safety} systematically evaluate DeepSeek models using CHiSafetyBench, revealing critical safety deficiencies specifically in Chinese contexts, where reasoning models like DeepSeek-R1 struggled with culturally-specific safety concerns and failed to adequately reject harmful prompts.

\subsection{Multi-modal Safety Risks}
\label{sec:multi-modal}

Following the success of LRMs, researchers have recognized the potential of reinforcement learning to enhance reasoning abilities in Large Vision-Language Models (LVLMs). This approach has led to the development of several notable models, including QvQ~\citep{qvq}, Mulberry~\citep{yao2024mulberryempoweringmllmo1like}, and R1-Onevision~\citep{yang2025r1onevisionadvancinggeneralizedmultimodal}. While these models demonstrate impressive reasoning capabilities, their safety implications remain largely unexplored. The pioneering work of SafeMLRM~\citep{fang2025safemlrmdemystifyingsafetymultimodal} provides the first systematic safety analysis of multi-modal large reasoning models, revealing three critical concerns: (1) acquiring reasoning capabilities significantly degrades inherited safety alignment, (2) certain scenarios exhibit disproportionately higher vulnerabilities, and (3) some models demonstrate nascent self-correction capabilities despite overall safety concerns. Given these findings, we emphasize the urgent need for comprehensive safety and vulnerability assessments of reasoning-enhanced LVLMs to ensure their responsible deployment and use.

\section{Attacks on LRMs}
\label{sec:attacks}
In this section, we categorize different attack methods based on their primary objectives. We identify four main categories: Reasoning Length Attacks (Section \ref{sec:reasoning_length}), which target the reasoning process itself; Answer Correctness Attacks (Section \ref{sec:answer_correctness}), which aim to manipulate output accuracy; Prompt Injection Attacks (Section \ref{sec:prompt_injection}), which bypass safety measures through crafted inputs; and Jailbreak Attacks (Section \ref{sec:jailbreak}), which attempt to extract prohibited content or behaviors. Each attack type exploits different vulnerabilities in the reasoning capabilities of LRMs.

\subsection{Reasoning Length Attacks}
\label{sec:reasoning_length}

Unlike traditional LLMs that generate direct responses, LRMs explicitly perform multi-step reasoning, creating a new attack surface related to reasoning length. Attackers can exploit this distinctive feature by either forcing models to overthink simple problems or short-cutting necessary deliberation processes.

\paragraph{Overthinking.}
The success of step-by-step reasoning in LRMs has significantly enhanced their problem-solving capabilities, but this improvement comes with a critical vulnerability: overthinking. Recent work by \citet{chen2024not} has identified that these models often spend orders of magnitude more computation on simple questions with minimal benefit, creating substantial inference overhead and latency issues. \citet{hashemi2025dnr} systematically demonstrate this inefficiency through their DNR benchmark, revealing that reasoning models generate up to 70× more tokens than necessary and often perform worse than simpler non-reasoning models on straightforward tasks. This inefficiency creates an exploitable attack surface where adversaries can deliberately trigger excessive reasoning through carefully crafted inputs. \citet{kumar2025overthinking} formalize this as an indirect prompt injection attack that introduces computationally demanding decoy problems, while \citet{zaremba2025tradinginferencetimecomputeadversarial} identify Nerd Sniping attacks that trap models in unproductive thinking loops, causing them to spend abnormally large amounts of inference-time compute with decreased performance. These attacks effectively apply denial-of-service techniques~\citep{9581273, gao2024denialofservicepoisoningattackslarge} specifically to LRMs. The implications extend beyond computational waste—\citet{marjanovic2025deepseek} and \citet{wu2025more} demonstrate that reasoning performance actually degrades beyond certain length thresholds, while \citet{cuadron2025dangeroverthinkingexaminingreasoningaction} show that in agentic systems, overthinking can lead to decision paralysis and ineffective action selection.

\paragraph{Underthinking.}
Complementing overthinking vulnerabilities, \citet{zaremba2025tradinginferencetimecomputeadversarial} propose Think Less attacks, where adversaries craft special prompts to force reasoning models to shortcut their deliberative processes. The goal is to make models produce incorrect responses by significantly reducing computation time. Their experiments use 64-shot examples to demonstrate that models like OpenAI's o1-mini are particularly susceptible to these attacks, bypassing normal reasoning and jumping to premature conclusions. However, this can be detected by monitoring for abnormally low inference-time compute usage.

\subsection{Answer Correctness Attacks}
\label{sec:answer_correctness}

While conventional LLMs can be manipulated to produce incorrect answers, LRMs introduce unique vulnerabilities through their exposed reasoning chains. This transparency in the inference process provides adversaries with additional attack vectors to corrupt the reasoning pathway itself, rather than just targeting the final output.

\paragraph{Reasoning-based Backdoor Attacks.}
The goal of backdoor attacks is to alter a model's behavior whenever a specific trigger is present in the input \citep{zhao2024survey}. Based on the nature of these triggers, backdoor attacks can be classified as instruction-based \citep{xu2023instructions}, prompt-based \citep{10446267}, or syntax-based \citep{qi2021hiddenkillerinvisibletextual, cheng2025synghostinvisibleuniversaltaskagnostic}.
With the advancement of reasoning capabilities in LRMs, a new paradigm has emerged: Chain-of-Thought (CoT) based backdoor attacks that specifically target intermediate reasoning steps to compromise answer correctness. BadChain \citep{xiang2024badchain} inserts malicious reasoning steps into the sequence, manipulating the model to produce incorrect answers while maintaining logical coherence. DarkMind \citep{guo2025darkmind} implements latent triggers that activate during specific reasoning scenarios, leading to plausible but false outputs that are difficult to detect. BoT \citep{zhu2025bot} forces models to bypass their reasoning mechanisms, generating immediate incorrect responses instead of thoughtful deliberation. ShadowCoT \citep{zhao2025shadowcot} directly manipulates the model's cognitive pathway through attention head localization and reasoning chain pollution, achieving flexible hijacking that produces wrong answers while preserving logical flow.
These sophisticated attacks reveal a concerning vulnerability: the enhanced reasoning capabilities of LRMs paradoxically make them more susceptible to backdoors that can generate incorrect answers accompanied by convincing reasoning.

\paragraph{Error Injection.}
The explicit reasoning processes of LRMs create a critical vulnerability where strategically injected errors can fundamentally compromise output integrity. \citet{cui2025processresultmanipulatedending} demonstrate this with their Compromising Thought (CPT) attack, where manipulating calculation results in reasoning tokens caused models to ignore correct steps and adopt incorrect answers. Their experiments with models like DeepSeek-R1 revealed that endpoint token manipulations had greater impact than structural changes to reasoning chains. They also discovered a security vulnerability where tampered tokens could trigger complete reasoning cessation in DeepSeek-R1, highlighting significant implications for reasoning-intensive applications.

\subsection{Prompt Injection Attacks}
\label{sec:prompt_injection}

Prompt injection attacks affect both traditional LLMs and LRMs, but LRMs present distinct challenges due to their step-by-step processing. These attacks~\citep{10555871,liu2023prompt, chen2024defense, chen2025can} inject malicious instructions disguised as normal user input, causing the AI to override or ignore its original developer-set instructions and safeguards. The explicit reasoning structures of LRMs offer attackers additional insertion points to redirect the model's thought process, potentially making them more susceptible to certain types of injections.

\citet{zhou2025hidden} examine LRMs like DeepSeek-R1 and o3-mini, finding significant differences in susceptibility based on injection types and risk categories. Their research reveals that reasoning models are particularly vulnerable to direct prompt injection attacks compared to indirect ones. \citet{zaremba2025tradinginferencetimecomputeadversarial} further demonstrate that open-source reasoning models show significant vulnerability to prompt injection attacks, with success rates varying between direct and indirect injections. Their experiments reveal that increasing inference-time compute substantially improves model robustness, with attack success probability decreasing as test-time compute grows. Notably, proprietary models like o3-mini demonstrate nearly 80\% lower vulnerability than open-source counterparts when facing direct injection attacks.

\subsection{Jailbreak Attacks}
\label{sec:jailbreak}

Jailbreak attacks~\citep{jin2024jailbreakzoosurveylandscapeshorizons,yi2024jailbreakattacksdefenseslarge} refer to methods designed to circumvent an AI system's safety guidelines and content policies to extract prohibited responses. While both traditional LLMs and LRMs face jailbreak threats, the attacks against LRMs represent a distinct category that specifically targets their enhanced reasoning capabilities. Rather than merely extending approaches used against conventional LLMs, these attacks exploit the deliberative processes that make LRMs powerful, enabling attackers to develop more sophisticated methods to bypass safety measures and elicit harmful content.

\paragraph{Prompt-Based Jailbreak.}
Prompt-based jailbreaks involve the careful crafting of prompts, employing techniques such as persuasion~\citep{zeng2024johnny}, nested scene construction~\citep{li2023deepinception}, and persona modulation~\citep{shah2023scalable}. \citet{andriushchenko2024does} introduce a method that applies past-tense transformations to OpenAI's recent o1 reasoning models, revealing their lack of robustness against subtle linguistic shifts. \citet{ying2025towards} propose attack prompts that combine common jailbreak strategies—such as scenario injection, affirmative prefixes, and indirect instructions—with safety-sensitive queries to probe model vulnerabilities. Their findings indicate that reasoning models like DeepSeek-R1 and OpenAI’s o1 are particularly susceptible to such attacks, as their explicit CoT reasoning renders them more exploitable than standard LLMs.

\paragraph{Multi-turn Jailbreak.}
Performing jailbreak attacks in a single query can be challenging, but multi-turn conversations or sequential prompts may incrementally guide models toward generating restricted content~\citet{russinovich2024crescendo, sun2024multi}. Multi-turn attacks are particularly relevant to reasoning-capable models as these models possess sophisticated logical processing that can be exploited through extended dialogues. \citet{ying2025reasoning} propose Reasoning-Augmented Conversation (RACE), which reformulates harmful queries into benign reasoning tasks and gradually exploits the model's inference capabilities to compromise safety alignment, achieving success rates up to 96\%. \citet{ren2024derail} introduce ActorAttack, a framework that constructs semantically linked conversational sequences that appear harmless individually but collectively lead to harmful outputs, successfully targeting even advanced models like o1. \citet{li2024llm} further show that multi-turn human jailbreaks significantly outperform automated single-turn attacks, leveraging the model's ability to maintain context and be incrementally steered toward unsafe behaviors.

\paragraph{Reasoning Exploitation Jailbreak.}
LRMs possess advanced reasoning capabilities that, while enhancing their utility, introduce unique vulnerabilities that can be exploited through reasoning-based jailbreak attacks. Unlike traditional LLMs, these models explicitly expose their CoT reasoning processes, creating new attack surfaces. ~\citet{yao2025mousetrap} introduce Mousetrap, a framework that leverages chaos mappings to create iterative reasoning chains that gradually lead LRMs into harmful outputs. By embedding one-to-one mappings into the reasoning process, Mousetrap effectively traps models like OpenAI's o1-mini and Claude-sonnet with success rates of up to 98\%. \citet{kuo2025h} propose Hijacking Chain-of-Thought (H-CoT), which manipulates the reasoning process by injecting execution-phase thoughts that bypass safety checks entirely. Their approach exploits LRMs' tendency to prioritize problem-solving over safety considerations, causing rejection rates to plummet from 98\% to below 2\% across models like OpenAI o1/o3 and DeepSeek-R1. Both approaches demonstrate that the very reasoning mechanisms designed to enhance LRMs' capabilities can become their most significant security weaknesses when strategically manipulated.

\section{Defenses for LRMs}
\label{sec:defenses}

To mitigate safety risks and defend against attacks on LRMs, various defense strategies have been proposed in recent research. We categorize these approaches into three main types: Safety Alignment (Section \ref{sec:safety_alignment}), Inference-Time Defenses (Section \ref{sec:inference_time_defenses}), and Guard Models (Section \ref{sec:guardrail}).

\subsection{Safety Alignment of LRMs}
\label{sec:safety_alignment}

Similar to LLMs and VLMs, LRMs are required to align with humans' values and expectations. The 3H principle \citep{askell2021general} (Helpful, Honest, and Harmless) provides a foundational guideline for constraining model behaviors.

The existing safety alignment pipelines and techniques developed for LLMs \citep{shen2023large} and VLMs \citep{vlm_safety_survey} can be readily adapted to LRMs, as they share similar architectures and natural language generation behaviors. For example, the alignment process for LLMs typically starts with collecting high-quality, value-aligned data \citep{ethayarajh2022understanding}, either from existing benchmarks \citep{bach2022promptsource,wang2022super}, LLM-generated instructions \citep{wang2022self}, or by filtering unsafe content \citep{welbl2021challenges,wang2022exploring}. During training, common techniques include supervised fine-tuning (SFT) \citep{sft}, reinforcement learning from human feedback (RLHF) \citep{rlhf}, and direct preference optimization (DPO) \citep{DPO}. In the domain of VLMs, safety alignment has been achieved through various approaches. For example, \citet{liu2024safety} introduce additional safety modules during training to enhance model alignment. Moreover, methods such as ADPO \citep{ADPO}, Safe RLHF-V \citep{safe_rlhf_v}, and GRPO-based methods \citep{grpo_safety} improve safety via DPO \citep{DPO}, RLHF \citep{rlhf}, and GRPO \citep{deepseekr1}, respectively. Additionally, open-source datasets and benchmarks \citep{zhang2024spa,safe_rlhf_v} have played a crucial role in providing high-quality alignment data for safety evaluation. 

Although effective, the previous alignment methods for LLMs and VLMs may overlook the reasoning process of LRMs, leading to unsatisfactory alignment performance. To mitigate this challenge, various works focus on different aspects, including safe CoT data curation, SFT-based safety alignment on reasoning, and RL-based safety alignment on reasoning.

\paragraph{Safe CoT Data Curation.} First, \citet{wang2025star} build a 1k-scale safety dataset named STAR-1 specifically designed for LRMs. Another safety training data in CoT style named SafeChain \citep{jiang2025safechain} is introduced to enhance the safety of LRMs. In addition, \citet{zhang2025realsafe} construct a dataset consisting of 15k safety-aware reasoning trajectories, generated by DeepSeek-R1, with explicit instructions designed to promote expected refusal behavior.

\paragraph{SFT-based Safety Alignment on Reasoning.}
Based on the curated safe CoT data, researchers further conduct SFT to improve safety. For example, \citet{jiang2025safechain} train two LRMs with the SafeChain dataset, demonstrating that it not only enhances model safety but also preserves reasoning performance. Besides, RealSafe-R1 \citep{zhang2025realsafe} is developed to make LRMs safer by training DeepSeek-R1 distilled models on the 15k safety-aware reasoning trajectories. \citet{wang2025leveraging} proposes training the model to reason with the guidelines, thereby enhancing survey alignment.

\paragraph{RL-based Safety Alignment on Reasoning.}
In addition to SFT, various further post-training techniques for safety are proposed based on reinforcement learning (RL). For example, deliberative alignment \citep{deliberative_alignment} is proposed to teach models safety specifications directly and train them to reason over these guidelines before generating responses explicitly via reinforcement learning. In addition, STAIR \citep{zhang2025stair} utilizes Monte Carlo tree search and DPO \citep{DPO} to integrate safety alignment with introspective reasoning. Meanwhile, a SaRO \citep{mou2025saro} is proposed to incorporate safety-policy-driven reasoning into the alignment process. Besides, R2D \citep{zhu2025reasoning} is present to unlock the safety-aware reasoning mechanism to defense against jailbreak attacks with the proposed contrastive pivot optimization (CPO). 

However, safety alignment brings the safety alignment tax \citep{lin2023mitigating}, compromising the fundamental capabilities of LRMs like reasoning capability \citep{huang2025safety}. To mitigate this issue, researchers explore alternative defense techniques that do not require direct modifications to the victim models.

\subsection{Inference-time Defenses for LRMs}
\label{sec:inference_time_defenses}
To circumvent the safety alignment tax \citep{lin2023mitigating,huang2025safety}, one line of work focuses on applying defenses at inference time. 
The insights from previous inference-time defenses for LLMs \citep{cheng2023black,lu2023inference} and VLMs \citep{wang2024inferaligner,ghosal2024immune,ding2024eta,vlm_guard}, such as safe system prompting, few-shot safe demonstrations, and safe decoding, can be naturally borrowed to LRMs, as the token generation mechanism is similar across these models.

However, the reasoning process in LRMs brings new challenges and opportunities for inference-time defenses. Therefore, various inference-time techniques like inference-time scaling on reasoning and safe decoding for reasoning are proposed to ensure the safety of reasoning in LRMs. 

\paragraph{Inference-time Scaling on Reasoning.}  \citet{zaremba2025tradinginferencetimecomputeadversarial} demonstrate that the inference-time scaling on reasoning improves the safety and adversarial robustness of LRMs. Future work could explore dynamic scaling strategies tailored to input complexity, or integrate adaptive reasoning depth control to balance efficiency and safety performance \citep{liu2025efficient} during inference.

\paragraph{Safe Decoding for Reasoning.} \citet{jiang2025safechain} propose three decoding strategies, including ZeroThink, LessThink, and MoreThink, to verify model safety during reasoning. Making the reasoning safer at inference time could be a promising future direction, by verifying intermediate steps, filtering unsafe trajectories, or integrating reasoning-aware guard mechanisms during decoding.~\citet{wu2025effectivelycontrollingreasoningmodels} introduce Thinking Intervention, a method that strategically injects guidance directly into the reasoning process to control model behavior and improve safety alignment without requiring additional training.

\subsection{Guard Models for LRMs}
\label{sec:guardrail}
Another line of work without direct modification to the victim model focuses on building guard models for the victim model. The previous inference-time defenses still focus on the safer inference of the victim models themselves. Differently, guard models aim to moderate the input and output of the victim models without training the victim models or modifying the inference strategies of the victim models. The existing guard models for LLMs \citep{Llamaguard} or VLMs \citep{llamaguard3_vision} can also safeguard the LRMs since they share similar input and output formats. In addition, the reasoning-based guard models \citep{GuardReasoner} can better moderate the reasoning process of LRMs via guiding the guard models to deliberatively reason before making moderation decisions. We category existing guard models into two classes, including classifier-based guard models and reasoning-based guard models.

\paragraph{Classifier-based Guard Models.} 
The LLM guard models, including ToxicChat-T5 \citep{Toxicchat}, ToxDectRoberta \citep{tocxic_roberta}, LaGoNN \citep{LaGoNN}, the LLaMA Guard series \citep{Llamaguard,llama3}, Aegis Guard series \citep{AegisGuard,AegisGuard2}, WildGuard \citep{wildguard}, ShieldGemma \citep{Shieldgemma}, are typically based on open-sourced LLMs and fine-tuned on the red-teaming data. In the VLM domain, for example, LLaVAGuard \citep{llava_guard} is built to conduct large-scale dataset annotation and moderate the text-image models. In addition, VLMGuard \citep{vlmguard} is proposed to conduct malicious image-text prompt detection by leveraging the unlabeled user prompts. Moreover, LLaMA Guard 3-Vision \citep{llama_guard_3_vision} is developed to moderate both the image-text input and text output of VLMs via SFT. To improve the generalization ability, \citep{safe_rlhf_v} presents Beaver-Guard-V by training a reward model and then applying reinforcement learning. Although effective, they are typically classifier-based guard models, limiting their abilities in moderate reasoning data. To mitigate this problem, the reasoning-based guard models \citep{GuardReasoner} are proposed to enhance the reasoning ability of guard models.

\paragraph{Reasoning-based Guard Models.}

Through the proposed reasoning SFT and hard sample DPO, GuardReasoner \citep{GuardReasoner} and GuardReasoner-VL~\citep{guardreasoner-vl} are proposed to guide the guard model to deliberatively reason before making moderation decisions, improving performance, generalization ability, and explainability. Similarly, ThinkGuard \citep{ThinkGuard} is developed via the proposed critique-augmented fine-tuning. X-Guard \citep{upadhayay2025x} extends the reasoning-based guard model to the multi-lingual scenario.

\section{Future Directions}
\label{sec:future}
Beyond the detailed analysis of risks, attacks, and defenses presented in previous sections, this paper also identifies future directions that researchers should prioritize to enhance the safety of LRMs: 
\textbf{(1) Standardized Evaluation Benchmarks.}
New benchmarks should focus on reasoning-specific vulnerabilities, as the research community currently lacks standardized evaluation frameworks to comprehensively test both the safety and robustness of LRMs' multi-step reasoning processes. \textbf{(2) Domain‐Specific Evaluation Frameworks.}
Evaluation suites for healthcare, finance, and law must include curated case studies and targeted adversarial tests. Expert review ensures LRMs meet each domain’s accuracy and ethical requirements. \textbf{(3) Human‐in‐the‐Loop Alignment and Interpretability.} Interactive tools should let experts inspect and refine reasoning traces. Iterative feedback can align LRMs with stakeholder values and correct biases efficiently.  

\section{Conclusion}
\label{sec:conclusion}
This survey has comprehensively examined the emerging safety challenges posed by Large Reasoning Models. Through our analysis, we've identified several critical insights that distinguish LRM safety from traditional LLMs. First, \textbf{LRMs expose their reasoning chains, creating new attack surfaces} where adversaries can manipulate intermediate steps rather than just outputs, enabling sophisticated attacks like reasoning-based backdoors and hijacking that target the deliberative process itself. Second, \textbf{traditional output-focused alignment methods prove insufficient for LRMs}, as harmful reasoning can persist internally even when final outputs appear safe, necessitating novel approaches that consider the entire reasoning trajectory. These insights underscore the need for specialized safety research targeting LRMs, including standardized evaluation benchmarks for reasoning-specific vulnerabilities and human-in-the-loop alignment methods that can inspect and refine reasoning traces as these powerful models continue to advance into increasingly critical domains.

\section*{Limitations}
This survey has inherent limitations due to the rapidly evolving nature of LRMs. Since the emergence of OpenAI's o1 series, DeepSeek-R1, and other advanced reasoning models is relatively recent, our taxonomy and findings may become outdated as new research continuously emerges. While we have endeavored to provide a comprehensive overview of safety challenges, attacks, and defenses, we acknowledge that some aspects may require revision as the field matures. Additionally, our reliance on published academic literature may not fully capture proprietary research being conducted within companies developing these models, potentially creating gaps in understanding industry-specific safety measures.

\bibliographystyle{acl_natbib}
\bibliography{custom}

\begin{thebibliography}{130}
\expandafter\ifx\csname natexlab\endcsname\relax\def\natexlab#1{#1}\fi

\bibitem[{Andriushchenko and Flammarion(2024)}]{andriushchenko2024does}
Maksym Andriushchenko and Nicolas Flammarion. 2024.
\newblock Does refusal training in llms generalize to the past tense?
\newblock \emph{arXiv preprint arXiv:2407.11969}.

\bibitem[{Arrieta et~al.(2025{\natexlab{a}})Arrieta, Ugarte, Valle, Parejo, and Segura}]{arrieta2025early}
Aitor Arrieta, Miriam Ugarte, Pablo Valle, Jos{\'e}~Antonio Parejo, and Sergio Segura. 2025{\natexlab{a}}.
\newblock Early external safety testing of openai's o3-mini: Insights from the pre-deployment evaluation.
\newblock \emph{arXiv preprint arXiv:2501.17749}.

\bibitem[{Arrieta et~al.(2025{\natexlab{b}})Arrieta, Ugarte, Valle, Parejo, and Segura}]{arrieta2025o3}
Aitor Arrieta, Miriam Ugarte, Pablo Valle, Jos{\'e}~Antonio Parejo, and Sergio Segura. 2025{\natexlab{b}}.
\newblock o3-mini vs deepseek-r1: Which one is safer?
\newblock \emph{arXiv preprint arXiv:2501.18438}.

\bibitem[{Askell et~al.(2021)Askell, Bai, Chen, Drain, Ganguli, Henighan, Jones, Joseph, Mann, DasSarma et~al.}]{askell2021general}
Amanda Askell, Yuntao Bai, Anna Chen, Dawn Drain, Deep Ganguli, Tom Henighan, Andy Jones, Nicholas Joseph, Ben Mann, Nova DasSarma, et~al. 2021.
\newblock A general language assistant as a laboratory for alignment.
\newblock \emph{arXiv preprint arXiv:2112.00861}.

\bibitem[{Bach et~al.(2022)Bach, Sanh, Yong, Webson, Raffel, Nayak, Sharma, Kim, Bari, Fevry et~al.}]{bach2022promptsource}
Stephen~H Bach, Victor Sanh, Zheng-Xin Yong, Albert Webson, Colin Raffel, Nihal~V Nayak, Abheesht Sharma, Taewoon Kim, M~Saiful Bari, Thibault Fevry, et~al. 2022.
\newblock Promptsource: An integrated development environment and repository for natural language prompts.
\newblock \emph{arXiv preprint arXiv:2202.01279}.

\bibitem[{Barkur et~al.(2025)Barkur, Schacht, and Scholl}]{barkur2025deception}
Sudarshan~Kamath Barkur, Sigurd Schacht, and Johannes Scholl. 2025.
\newblock Deception in llms: Self-preservation and autonomous goals in large language models.
\newblock \emph{arXiv preprint arXiv:2501.16513}.

\bibitem[{Bates and Gurevych(2023)}]{LaGoNN}
Luke Bates and Iryna Gurevych. 2023.
\newblock Like a good nearest neighbor: Practical content moderation and text classification.
\newblock \emph{arXiv preprint arXiv:2302.08957}.

\bibitem[{Besta et~al.(2024)Besta, Blach, Kubicek, Gerstenberger, Podstawski, Gianinazzi, Gajda, Lehmann, Niewiadomski, Nyczyk, and Hoefler}]{Besta_2024}
Maciej Besta, Nils Blach, Ales Kubicek, Robert Gerstenberger, Michal Podstawski, Lukas Gianinazzi, Joanna Gajda, Tomasz Lehmann, Hubert Niewiadomski, Piotr Nyczyk, and Torsten Hoefler. 2024.
\newblock \href {https://doi.org/10.1609/aaai.v38i16.29720} {Graph of thoughts: Solving elaborate problems with large language models}.
\newblock \emph{Proceedings of the AAAI Conference on Artificial Intelligence}, 38(16):17682–17690.

\bibitem[{Bondarenko et~al.(2025)Bondarenko, Volk, Volkov, and Ladish}]{bondarenko2025demonstrating}
Alexander Bondarenko, Denis Volk, Dmitrii Volkov, and Jeffrey Ladish. 2025.
\newblock Demonstrating specification gaming in reasoning models.
\newblock \emph{arXiv preprint arXiv:2502.13295}.

\bibitem[{Chen et~al.(2024{\natexlab{a}})Chen, Xu, Liang, He, Pang, Yu, Song, Liu, Zhou, Zhang et~al.}]{chen2024not}
Xingyu Chen, Jiahao Xu, Tian Liang, Zhiwei He, Jianhui Pang, Dian Yu, Linfeng Song, Qiuzhi Liu, Mengfei Zhou, Zhuosheng Zhang, et~al. 2024{\natexlab{a}}.
\newblock Do not think that much for 2+ 3=? on the overthinking of o1-like llms.
\newblock \emph{arXiv preprint arXiv:2412.21187}.

\bibitem[{Chen et~al.(2025)Chen, Li, Sui, He, Liu, Song, and Hooi}]{chen2025can}
Yulin Chen, Haoran Li, Yuan Sui, Yufei He, Yue Liu, Yangqiu Song, and Bryan Hooi. 2025.
\newblock Can indirect prompt injection attacks be detected and removed?
\newblock \emph{arXiv preprint arXiv:2502.16580}.

\bibitem[{Chen et~al.(2024{\natexlab{b}})Chen, Li, Zheng, Song, Wu, and Hooi}]{chen2024defense}
Yulin Chen, Haoran Li, Zihao Zheng, Yangqiu Song, Dekai Wu, and Bryan Hooi. 2024{\natexlab{b}}.
\newblock Defense against prompt injection attack by leveraging attack techniques.
\newblock \emph{arXiv preprint arXiv:2411.00459}.

\bibitem[{Cheng et~al.(2023)Cheng, Liu, Zheng, Ke, Wang, Dong, Tang, and Huang}]{cheng2023black}
Jiale Cheng, Xiao Liu, Kehan Zheng, Pei Ke, Hongning Wang, Yuxiao Dong, Jie Tang, and Minlie Huang. 2023.
\newblock Black-box prompt optimization: Aligning large language models without model training.
\newblock \emph{arXiv preprint arXiv:2311.04155}.

\bibitem[{Cheng et~al.(2025)Cheng, Du, Wu, Zhang, Chen, Zhang, and Liu}]{cheng2025synghostinvisibleuniversaltaskagnostic}
Pengzhou Cheng, Wei Du, Zongru Wu, Fengwei Zhang, Libo Chen, Zhuosheng Zhang, and Gongshen Liu. 2025.
\newblock \href {http://arxiv.org/abs/2402.18945} {Synghost: Invisible and universal task-agnostic backdoor attack via syntactic transfer}.

\bibitem[{Chi et~al.(2024{\natexlab{a}})Chi, Karn, Zhan, Smith, Rando, Zhang, Plawiak, Coudert, Upasani, and Pasupuleti}]{llama_guard_3_vision}
Jianfeng Chi, Ujjwal Karn, Hongyuan Zhan, Eric Smith, Javier Rando, Yiming Zhang, Kate Plawiak, Zacharie~Delpierre Coudert, Kartikeya Upasani, and Mahesh Pasupuleti. 2024{\natexlab{a}}.
\newblock Llama guard 3 vision: Safeguarding human-ai image understanding conversations.
\newblock \emph{arXiv preprint arXiv:2411.10414}.

\bibitem[{Chi et~al.(2024{\natexlab{b}})Chi, Karn, Zhan, Smith, Rando, Zhang, Plawiak, Coudert, Upasani, and Pasupuleti}]{llamaguard3_vision}
Jianfeng Chi, Ujjwal Karn, Hongyuan Zhan, Eric Smith, Javier Rando, Yiming Zhang, Kate Plawiak, Zacharie~Delpierre Coudert, Kartikeya Upasani, and Mahesh Pasupuleti. 2024{\natexlab{b}}.
\newblock Llama guard 3 vision: Safeguarding human-ai image understanding conversations.
\newblock \emph{arXiv preprint arXiv:2411.10414}.

\bibitem[{Cuadron et~al.(2025)Cuadron, Li, Ma, Wang, Wang, Zhuang, Liu, Schroeder, Xia, Mao, Thumiger, Desai, Stoica, Klimovic, Neubig, and Gonzalez}]{cuadron2025dangeroverthinkingexaminingreasoningaction}
Alejandro Cuadron, Dacheng Li, Wenjie Ma, Xingyao Wang, Yichuan Wang, Siyuan Zhuang, Shu Liu, Luis~Gaspar Schroeder, Tian Xia, Huanzhi Mao, Nicholas Thumiger, Aditya Desai, Ion Stoica, Ana Klimovic, Graham Neubig, and Joseph~E. Gonzalez. 2025.
\newblock \href {http://arxiv.org/abs/2502.08235} {The danger of overthinking: Examining the reasoning-action dilemma in agentic tasks}.

\bibitem[{Cui et~al.(2025)Cui, Hooi, Cai, and Wang}]{cui2025processresultmanipulatedending}
Yu~Cui, Bryan Hooi, Yujun Cai, and Yiwei Wang. 2025.
\newblock \href {http://arxiv.org/abs/2503.19326} {Process or result? manipulated ending tokens can mislead reasoning llms to ignore the correct reasoning steps}.

\bibitem[{DeepSeek-AI et~al.(2025)DeepSeek-AI, Guo, Yang, Zhang, Song, Zhang, Xu, Zhu, Ma, Wang, Bi, Zhang, Yu, Wu, Wu, Gou, Shao, Li, Gao, Liu, Xue, Wang, Wu, Feng, Lu, Zhao, Deng, Zhang, Ruan, Dai, Chen, Ji, Li, Lin, Dai, Luo, Hao, Chen, Li, Zhang, Bao, Xu, Wang, Ding, Xin, Gao, Qu, Li, Guo, Li, Wang, Chen, Yuan, Qiu, Li, Cai, Ni, Liang, Chen, Dong, Hu, Gao, Guan, Huang, Yu, Wang, Zhang, Zhao, Wang, Zhang, Xu, Xia, Zhang, Zhang, Tang, Li, Wang, Li, Tian, Huang, Zhang, Wang, Chen, Du, Ge, Zhang, Pan, Wang, Chen, Jin, Chen, Lu, Zhou, Chen, Ye, Wang, Yu, Zhou, Pan, Li, Zhou, Wu, Ye, Yun, Pei, Sun, Wang, Zeng, Zhao, Liu, Liang, Gao, Yu, Zhang, Xiao, An, Liu, Wang, Chen, Nie, Cheng, Liu, Xie, Liu, Yang, Li, Su, Lin, Li, Jin, Shen, Chen, Sun, Wang, Song, Zhou, Wang, Shan, Li, Wang, Wei, Zhang, Xu, Li, Zhao, Sun, Wang, Yu, Zhang, Shi, Xiong, He, Piao, Wang, Tan, Ma, Liu, Guo, Ou, Wang, Gong, Zou, He, Xiong, Luo, You, Liu, Zhou, Zhu, Xu, Huang, Li, Zheng, Zhu, Ma, Tang, Zha, Yan, Ren, Ren, Sha, Fu, Xu, Xie, Zhang,
  Hao, Ma, Yan, Wu, Gu, Zhu, Liu, Li, Xie, Song, Pan, Huang, Xu, Zhang, and Zhang}]{deepseekr1}
DeepSeek-AI, Daya Guo, Dejian Yang, Haowei Zhang, Junxiao Song, Ruoyu Zhang, Runxin Xu, Qihao Zhu, Shirong Ma, Peiyi Wang, Xiao Bi, Xiaokang Zhang, Xingkai Yu, Yu~Wu, Z.~F. Wu, Zhibin Gou, Zhihong Shao, Zhuoshu Li, Ziyi Gao, Aixin Liu, Bing Xue, Bingxuan Wang, Bochao Wu, Bei Feng, Chengda Lu, Chenggang Zhao, Chengqi Deng, Chenyu Zhang, Chong Ruan, Damai Dai, Deli Chen, Dongjie Ji, Erhang Li, Fangyun Lin, Fucong Dai, Fuli Luo, Guangbo Hao, Guanting Chen, Guowei Li, H.~Zhang, Han Bao, Hanwei Xu, Haocheng Wang, Honghui Ding, Huajian Xin, Huazuo Gao, Hui Qu, Hui Li, Jianzhong Guo, Jiashi Li, Jiawei Wang, Jingchang Chen, Jingyang Yuan, Junjie Qiu, Junlong Li, J.~L. Cai, Jiaqi Ni, Jian Liang, Jin Chen, Kai Dong, Kai Hu, Kaige Gao, Kang Guan, Kexin Huang, Kuai Yu, Lean Wang, Lecong Zhang, Liang Zhao, Litong Wang, Liyue Zhang, Lei Xu, Leyi Xia, Mingchuan Zhang, Minghua Zhang, Minghui Tang, Meng Li, Miaojun Wang, Mingming Li, Ning Tian, Panpan Huang, Peng Zhang, Qiancheng Wang, Qinyu Chen, Qiushi Du, Ruiqi Ge, Ruisong
  Zhang, Ruizhe Pan, Runji Wang, R.~J. Chen, R.~L. Jin, Ruyi Chen, Shanghao Lu, Shangyan Zhou, Shanhuang Chen, Shengfeng Ye, Shiyu Wang, Shuiping Yu, Shunfeng Zhou, Shuting Pan, S.~S. Li, Shuang Zhou, Shaoqing Wu, Shengfeng Ye, Tao Yun, Tian Pei, Tianyu Sun, T.~Wang, Wangding Zeng, Wanjia Zhao, Wen Liu, Wenfeng Liang, Wenjun Gao, Wenqin Yu, Wentao Zhang, W.~L. Xiao, Wei An, Xiaodong Liu, Xiaohan Wang, Xiaokang Chen, Xiaotao Nie, Xin Cheng, Xin Liu, Xin Xie, Xingchao Liu, Xinyu Yang, Xinyuan Li, Xuecheng Su, Xuheng Lin, X.~Q. Li, Xiangyue Jin, Xiaojin Shen, Xiaosha Chen, Xiaowen Sun, Xiaoxiang Wang, Xinnan Song, Xinyi Zhou, Xianzu Wang, Xinxia Shan, Y.~K. Li, Y.~Q. Wang, Y.~X. Wei, Yang Zhang, Yanhong Xu, Yao Li, Yao Zhao, Yaofeng Sun, Yaohui Wang, Yi~Yu, Yichao Zhang, Yifan Shi, Yiliang Xiong, Ying He, Yishi Piao, Yisong Wang, Yixuan Tan, Yiyang Ma, Yiyuan Liu, Yongqiang Guo, Yuan Ou, Yuduan Wang, Yue Gong, Yuheng Zou, Yujia He, Yunfan Xiong, Yuxiang Luo, Yuxiang You, Yuxuan Liu, Yuyang Zhou, Y.~X. Zhu,
  Yanhong Xu, Yanping Huang, Yaohui Li, Yi~Zheng, Yuchen Zhu, Yunxian Ma, Ying Tang, Yukun Zha, Yuting Yan, Z.~Z. Ren, Zehui Ren, Zhangli Sha, Zhe Fu, Zhean Xu, Zhenda Xie, Zhengyan Zhang, Zhewen Hao, Zhicheng Ma, Zhigang Yan, Zhiyu Wu, Zihui Gu, Zijia Zhu, Zijun Liu, Zilin Li, Ziwei Xie, Ziyang Song, Zizheng Pan, Zhen Huang, Zhipeng Xu, Zhongyu Zhang, and Zhen Zhang. 2025.
\newblock \href {http://arxiv.org/abs/2501.12948} {Deepseek-r1: Incentivizing reasoning capability in llms via reinforcement learning}.

\bibitem[{Ding et~al.(2024)Ding, Li, and Zhang}]{ding2024eta}
Yi~Ding, Bolian Li, and Ruqi Zhang. 2024.
\newblock Eta: Evaluating then aligning safety of vision language models at inference time.
\newblock \emph{arXiv preprint arXiv:2410.06625}.

\bibitem[{Du et~al.(2024)Du, Ghosh, Sim, Salem, Carvalho, Lawton, Li, and Stokes}]{vlmguard}
Xuefeng Du, Reshmi Ghosh, Robert Sim, Ahmed Salem, Vitor Carvalho, Emily Lawton, Yixuan Li, and Jack~W Stokes. 2024.
\newblock Vlmguard: Defending vlms against malicious prompts via unlabeled data.
\newblock \emph{arXiv preprint arXiv:2410.00296}.

\bibitem[{Dubey et~al.(2024)Dubey, Jauhri, Pandey, Kadian, Al-Dahle, Letman, Mathur, Schelten, Yang, Fan et~al.}]{llama3}
Abhimanyu Dubey, Abhinav Jauhri, Abhinav Pandey, Abhishek Kadian, Ahmad Al-Dahle, Aiesha Letman, Akhil Mathur, Alan Schelten, Amy Yang, Angela Fan, et~al. 2024.
\newblock The llama 3 herd of models.
\newblock \emph{arXiv preprint arXiv:2407.21783}.

\bibitem[{Ethayarajh et~al.(2022)Ethayarajh, Choi, and Swayamdipta}]{ethayarajh2022understanding}
Kawin Ethayarajh, Yejin Choi, and Swabha Swayamdipta. 2022.
\newblock Understanding dataset difficulty with mathcal v-usable information.
\newblock In \emph{International Conference on Machine Learning}. PMLR.

\bibitem[{Fang et~al.(2025)Fang, Wang, Wang, Yao, Wang, Zhang, Wang, and Chua}]{fang2025safemlrmdemystifyingsafetymultimodal}
Junfeng Fang, Yukai Wang, Ruipeng Wang, Zijun Yao, Kun Wang, An~Zhang, Xiang Wang, and Tat-Seng Chua. 2025.
\newblock \href {http://arxiv.org/abs/2504.08813} {Safemlrm: Demystifying safety in multi-modal large reasoning models}.

\bibitem[{Feng et~al.(2023)Feng, Wan, Wen, McAleer, Wen, Zhang, and Wang}]{alphazero}
Xidong Feng, Ziyu Wan, Muning Wen, Stephen~Marcus McAleer, Ying Wen, Weinan Zhang, and Jun Wang. 2023.
\newblock Alphazero-like tree-search can guide large language model decoding and training.
\newblock \emph{arXiv preprint arXiv:2309.17179}.

\bibitem[{Gao et~al.(2024)Gao, Pang, Du, Yang, Xia, and Lin}]{gao2024denialofservicepoisoningattackslarge}
Kuofeng Gao, Tianyu Pang, Chao Du, Yong Yang, Shu-Tao Xia, and Min Lin. 2024.
\newblock \href {http://arxiv.org/abs/2410.10760} {Denial-of-service poisoning attacks against large language models}.

\bibitem[{Ghosal et~al.(2024)Ghosal, Chakraborty, Singh, Guan, Wang, Beirami, Huang, Velasquez, Manocha, and Bedi}]{ghosal2024immune}
Soumya~Suvra Ghosal, Souradip Chakraborty, Vaibhav Singh, Tianrui Guan, Mengdi Wang, Ahmad Beirami, Furong Huang, Alvaro Velasquez, Dinesh Manocha, and Amrit~Singh Bedi. 2024.
\newblock Immune: Improving safety against jailbreaks in multi-modal llms via inference-time alignment.
\newblock \emph{arXiv preprint arXiv:2411.18688}.

\bibitem[{Ghosh et~al.(2024{\natexlab{a}})Ghosh, Varshney, Galinkin, and Parisien}]{AegisGuard}
Shaona Ghosh, Prasoon Varshney, Erick Galinkin, and Christopher Parisien. 2024{\natexlab{a}}.
\newblock Aegis: Online adaptive ai content safety moderation with ensemble of llm experts.
\newblock \emph{arXiv preprint arXiv:2404.05993}.

\bibitem[{Ghosh et~al.(2024{\natexlab{b}})Ghosh, Varshney, Sreedhar, Padmakumar, Rebedea, Varghese, and Parisien}]{AegisGuard2}
Shaona Ghosh, Prasoon Varshney, Makesh~Narsimhan Sreedhar, Aishwarya Padmakumar, Traian Rebedea, Jibin~Rajan Varghese, and Christopher Parisien. 2024{\natexlab{b}}.
\newblock Aegis2. 0: A diverse ai safety dataset and risks taxonomy for alignment of llm guardrails.
\newblock In \emph{Neurips Safe Generative AI Workshop 2024}.

\bibitem[{Guan et~al.(2024)Guan, Joglekar, Wallace, Jain, Barak, Heylar, Dias, Vallone, Ren, Wei et~al.}]{deliberative_alignment}
Melody~Y Guan, Manas Joglekar, Eric Wallace, Saachi Jain, Boaz Barak, Alec Heylar, Rachel Dias, Andrea Vallone, Hongyu Ren, Jason Wei, et~al. 2024.
\newblock Deliberative alignment: Reasoning enables safer language models.
\newblock \emph{arXiv preprint arXiv:2412.16339}.

\bibitem[{Guo and Tourani(2025)}]{guo2025darkmind}
Zhen Guo and Reza Tourani. 2025.
\newblock Darkmind: Latent chain-of-thought backdoor in customized llms.
\newblock \emph{arXiv preprint arXiv:2501.18617}.

\bibitem[{Han et~al.(2024)Han, Rao, Ettinger, Jiang, Lin, Lambert, Choi, and Dziri}]{wildguard}
Seungju Han, Kavel Rao, Allyson Ettinger, Liwei Jiang, Bill~Yuchen Lin, Nathan Lambert, Yejin Choi, and Nouha Dziri. 2024.
\newblock Wildguard: Open one-stop moderation tools for safety risks, jailbreaks, and refusals of llms.
\newblock \emph{arXiv preprint arXiv:2406.18495}.

\bibitem[{Hashemi et~al.(2025)Hashemi, Bamgbose, Madhusudhan, Nair, Tiwari, and Yadav}]{hashemi2025dnr}
Masoud Hashemi, Oluwanifemi Bamgbose, Sathwik~Tejaswi Madhusudhan, Jishnu~Sethumadhavan Nair, Aman Tiwari, and Vikas Yadav. 2025.
\newblock Dnr bench: When silence is smarter--benchmarking over-reasoning in reasoning llms.
\newblock \emph{arXiv preprint arXiv:2503.15793}.

\bibitem[{He et~al.(2025)He, Li, Wu, Sui, Chen, and Hooi}]{he2025evaluating}
Yufei He, Yuexin Li, Jiaying Wu, Yuan Sui, Yulin Chen, and Bryan Hooi. 2025.
\newblock Evaluating the paperclip maximizer: Are rl-based language models more likely to pursue instrumental goals?
\newblock \emph{arXiv preprint arXiv:2502.12206}.

\bibitem[{Helff et~al.(2024)Helff, Friedrich, Brack, Schramowski, and Kersting}]{llava_guard}
Lukas Helff, Felix Friedrich, Manuel Brack, Patrick Schramowski, and Kristian Kersting. 2024.
\newblock Llavaguard: Vlm-based safeguard for vision dataset curation and safety assessment.
\newblock In \emph{Proceedings of the IEEE/CVF Conference on Computer Vision and Pattern Recognition}, pages 8322--8326.

\bibitem[{Huang et~al.(2025)Huang, Hu, Ilhan, Tekin, Yahn, Xu, and Liu}]{huang2025safety}
Tiansheng Huang, Sihao Hu, Fatih Ilhan, Selim~Furkan Tekin, Zachary Yahn, Yichang Xu, and Ling Liu. 2025.
\newblock Safety tax: Safety alignment makes your large reasoning models less reasonable.
\newblock \emph{arXiv preprint arXiv:2503.00555}.

\bibitem[{Huang et~al.(2023)Huang, Ruan, Huang, Jin, Dong, Wu, Bensalem, Mu, Qi, Zhao, Cai, Zhang, Wu, Xu, Wu, Freitas, and Mustafa}]{huang2023surveysafetytrustworthinesslarge}
Xiaowei Huang, Wenjie Ruan, Wei Huang, Gaojie Jin, Yi~Dong, Changshun Wu, Saddek Bensalem, Ronghui Mu, Yi~Qi, Xingyu Zhao, Kaiwen Cai, Yanghao Zhang, Sihao Wu, Peipei Xu, Dengyu Wu, Andre Freitas, and Mustafa~A. Mustafa. 2023.
\newblock \href {http://arxiv.org/abs/2305.11391} {A survey of safety and trustworthiness of large language models through the lens of verification and validation}.

\bibitem[{Inan et~al.(2023)Inan, Upasani, Chi, Rungta, Iyer, Mao, Tontchev, Hu, Fuller, Testuggine et~al.}]{Llamaguard}
Hakan Inan, Kartikeya Upasani, Jianfeng Chi, Rashi Rungta, Krithika Iyer, Yuning Mao, Michael Tontchev, Qing Hu, Brian Fuller, Davide Testuggine, et~al. 2023.
\newblock Llama guard: Llm-based input-output safeguard for human-ai conversations.
\newblock \emph{arXiv preprint arXiv:2312.06674}.

\bibitem[{Jain et~al.(2024)Jain, Han, Gu, Li, Yan, Zhang, Wang, Solar-Lezama, Sen, and Stoica}]{jain2024livecodebench}
Naman Jain, King Han, Alex Gu, Wen-Ding Li, Fanjia Yan, Tianjun Zhang, Sida Wang, Armando Solar-Lezama, Koushik Sen, and Ion Stoica. 2024.
\newblock Livecodebench: Holistic and contamination free evaluation of large language models for code.
\newblock \emph{arXiv preprint arXiv:2403.07974}.

\bibitem[{Ji et~al.(2025)Ji, Chen, Pan, Zhu, Zhang, Li, Hong, Chen, Zhou, Wang et~al.}]{safe_rlhf_v}
Jiaming Ji, Xinyu Chen, Rui Pan, Han Zhu, Conghui Zhang, Jiahao Li, Donghai Hong, Boyuan Chen, Jiayi Zhou, Kaile Wang, et~al. 2025.
\newblock Safe rlhf-v: Safe reinforcement learning from human feedback in multimodal large language models.
\newblock \emph{arXiv preprint arXiv:2503.17682}.

\bibitem[{Jiang et~al.(2025)Jiang, Xu, Li, Niu, Xiang, Li, Lin, and Poovendran}]{jiang2025safechain}
Fengqing Jiang, Zhangchen Xu, Yuetai Li, Luyao Niu, Zhen Xiang, Bo~Li, Bill~Yuchen Lin, and Radha Poovendran. 2025.
\newblock Safechain: Safety of language models with long chain-of-thought reasoning capabilities.
\newblock \emph{arXiv preprint arXiv:2502.12025}.

\bibitem[{Jin et~al.(2024)Jin, Hu, Li, Zhang, Chen, Zhuang, and Wang}]{jin2024jailbreakzoosurveylandscapeshorizons}
Haibo Jin, Leyang Hu, Xinuo Li, Peiyan Zhang, Chonghan Chen, Jun Zhuang, and Haohan Wang. 2024.
\newblock \href {http://arxiv.org/abs/2407.01599} {Jailbreakzoo: Survey, landscapes, and horizons in jailbreaking large language and vision-language models}.

\bibitem[{Ke et~al.(2023)Ke, Wen, Feng, Liu, Lei, Cheng, Wang, Zeng, Dong, Wang et~al.}]{ke2023critiquellm}
Pei Ke, Bosi Wen, Zhuoer Feng, Xiao Liu, Xuanyu Lei, Jiale Cheng, Shengyuan Wang, Aohan Zeng, Yuxiao Dong, Hongning Wang, et~al. 2023.
\newblock Critiquellm: Towards an informative critique generation model for evaluation of large language model generation.
\newblock \emph{arXiv preprint arXiv:2311.18702}.

\bibitem[{Ke et~al.(2025)Ke, Jiao, Ming, Nguyen, Xu, Long, Li, Qin, Wang, Savarese, Xiong, and Joty}]{ke2025surveyfrontiersllmreasoning}
Zixuan Ke, Fangkai Jiao, Yifei Ming, Xuan-Phi Nguyen, Austin Xu, Do~Xuan Long, Minzhi Li, Chengwei Qin, Peifeng Wang, Silvio Savarese, Caiming Xiong, and Shafiq Joty. 2025.
\newblock \href {http://arxiv.org/abs/2504.09037} {A survey of frontiers in llm reasoning: Inference scaling, learning to reason, and agentic systems}.

\bibitem[{Kojima et~al.(2022)Kojima, Gu, Reid, Matsuo, and Iwasawa}]{kojima2022large}
Takeshi Kojima, Shixiang~Shane Gu, Machel Reid, Yutaka Matsuo, and Yusuke Iwasawa. 2022.
\newblock Large language models are zero-shot reasoners.
\newblock \emph{Advances in neural information processing systems}, 35:22199--22213.

\bibitem[{Kumar et~al.(2025)Kumar, Roh, Naseh, Karpinska, Iyyer, Houmansadr, and Bagdasarian}]{kumar2025overthinking}
Abhinav Kumar, Jaechul Roh, Ali Naseh, Marzena Karpinska, Mohit Iyyer, Amir Houmansadr, and Eugene Bagdasarian. 2025.
\newblock Overthinking: Slowdown attacks on reasoning llms.
\newblock \emph{arXiv preprint arXiv:2502.02542}.

\bibitem[{Kumar et~al.(2024)Kumar, Cummings, and Stimpson}]{10555871}
Surender~Suresh Kumar, M.L. Cummings, and Alexander Stimpson. 2024.
\newblock \href {https://doi.org/10.1109/ICHMS59971.2024.10555871} {Strengthening llm trust boundaries: A survey of prompt injection attacks surender suresh kumar dr. m.l. cummings dr. alexander stimpson}.
\newblock In \emph{2024 IEEE 4th International Conference on Human-Machine Systems (ICHMS)}, pages 1--6.

\bibitem[{Kuo et~al.(2025)Kuo, Zhang, Ding, Wang, DiValentin, Bao, Wei, Li, and Chen}]{kuo2025h}
Martin Kuo, Jianyi Zhang, Aolin Ding, Qinsi Wang, Louis DiValentin, Yujia Bao, Wei Wei, Hai Li, and Yiran Chen. 2025.
\newblock H-cot: Hijacking the chain-of-thought safety reasoning mechanism to jailbreak large reasoning models, including openai o1/o3, deepseek-r1, and gemini 2.0 flash thinking.
\newblock \emph{arXiv preprint arXiv:2502.12893}.

\bibitem[{Li et~al.(2024)Li, Han, Steneker, Primack, Goodside, Zhang, Wang, Menghini, and Yue}]{li2024llm}
Nathaniel Li, Ziwen Han, Ian Steneker, Willow Primack, Riley Goodside, Hugh Zhang, Zifan Wang, Cristina Menghini, and Summer Yue. 2024.
\newblock Llm defenses are not robust to multi-turn human jailbreaks yet.
\newblock \emph{arXiv preprint arXiv:2408.15221}.

\bibitem[{Li et~al.(2023)Li, Zhou, Zhu, Yao, Liu, and Han}]{li2023deepinception}
Xuan Li, Zhanke Zhou, Jianing Zhu, Jiangchao Yao, Tongliang Liu, and Bo~Han. 2023.
\newblock Deepinception: Hypnotize large language model to be jailbreaker.
\newblock \emph{arXiv preprint arXiv:2311.03191}.

\bibitem[{Li et~al.(2025{\natexlab{a}})Li, Li, Kosuga, and Bian}]{grpo_safety}
Xuying Li, Zhuo Li, Yuji Kosuga, and Victor Bian. 2025{\natexlab{a}}.
\newblock Optimizing safe and aligned language generation: A multi-objective grpo approach.
\newblock \emph{arXiv preprint arXiv:2503.21819}.

\bibitem[{Li et~al.(2025{\natexlab{b}})Li, Zhang, Zhang, Zhang, Liu, Yao, Xu, Zheng, Wang, Chen et~al.}]{system_1_to_2}
Zhong-Zhi Li, Duzhen Zhang, Ming-Liang Zhang, Jiaxin Zhang, Zengyan Liu, Yuxuan Yao, Haotian Xu, Junhao Zheng, Pei-Jie Wang, Xiuyi Chen, et~al. 2025{\natexlab{b}}.
\newblock From system 1 to system 2: A survey of reasoning large language models.
\newblock \emph{arXiv preprint arXiv:2502.17419}.

\bibitem[{Liang et~al.(2023)Liang, He, Jiao, Wang, Wang, Wang, Yang, Shi, and Tu}]{liang2023encouraging}
Tian Liang, Zhiwei He, Wenxiang Jiao, Xing Wang, Yan Wang, Rui Wang, Yujiu Yang, Shuming Shi, and Zhaopeng Tu. 2023.
\newblock Encouraging divergent thinking in large language models through multi-agent debate.
\newblock \emph{arXiv preprint arXiv:2305.19118}.

\bibitem[{Lightman et~al.(2023)Lightman, Kosaraju, Burda, Edwards, Baker, Lee, Leike, Schulman, Sutskever, and Cobbe}]{lightman2023let}
Hunter Lightman, Vineet Kosaraju, Yuri Burda, Harrison Edwards, Bowen Baker, Teddy Lee, Jan Leike, John Schulman, Ilya Sutskever, and Karl Cobbe. 2023.
\newblock Let's verify step by step.
\newblock In \emph{The Twelfth International Conference on Learning Representations}.

\bibitem[{Lin et~al.(2023{\natexlab{a}})Lin, Lin, Xiong, Diao, Liu, Zhang, Pan, Wang, Hu, Zhang et~al.}]{lin2023mitigating}
Yong Lin, Hangyu Lin, Wei Xiong, Shizhe Diao, Jianmeng Liu, Jipeng Zhang, Rui Pan, Haoxiang Wang, Wenbin Hu, Hanning Zhang, et~al. 2023{\natexlab{a}}.
\newblock Mitigating the alignment tax of rlhf.
\newblock \emph{arXiv preprint arXiv:2309.06256}.

\bibitem[{Lin et~al.(2023{\natexlab{b}})Lin, Wang, Tong, Wang, Guo, Wang, and Shang}]{Toxicchat}
Zi~Lin, Zihan Wang, Yongqi Tong, Yangkun Wang, Yuxin Guo, Yujia Wang, and Jingbo Shang. 2023{\natexlab{b}}.
\newblock Toxicchat: Unveiling hidden challenges of toxicity detection in real-world user-ai conversation.
\newblock \emph{arXiv preprint arXiv:2310.17389}.

\bibitem[{Liu et~al.(2025{\natexlab{a}})Liu, Wang, Xiao, and Chen}]{vlm_guard}
Qin Liu, Fei Wang, Chaowei Xiao, and Muhao Chen. 2025{\natexlab{a}}.
\newblock Vlm-guard: Safeguarding vision-language models via fulfilling safety alignment gap.
\newblock \emph{arXiv preprint arXiv:2502.10486}.

\bibitem[{Liu et~al.(2023)Liu, Deng, Li, Wang, Wang, Wang, Zhang, Liu, Wang, Zheng et~al.}]{liu2023prompt}
Yi~Liu, Gelei Deng, Yuekang Li, Kailong Wang, Zihao Wang, Xiaofeng Wang, Tianwei Zhang, Yepang Liu, Haoyu Wang, Yan Zheng, et~al. 2023.
\newblock Prompt injection attack against llm-integrated applications.
\newblock \emph{arXiv preprint arXiv:2306.05499}.

\bibitem[{Liu et~al.(2025{\natexlab{b}})Liu, Gao, Zhai, Jun, Wu, Xue, Chen, Kawaguchi, Zhang, and Hooi}]{GuardReasoner}
Yue Liu, Hongcheng Gao, Shengfang Zhai, Xia Jun, Tianyi Wu, Zhiwei Xue, Yulin Chen, Kenji Kawaguchi, Jiaheng Zhang, and Bryan Hooi. 2025{\natexlab{b}}.
\newblock Guardreasoner: Towards reasoning-based llm safeguards.
\newblock \emph{arXiv preprint arXiv:2501.18492}.

\bibitem[{Liu et~al.(2025{\natexlab{c}})Liu, Wu, He, Gao, Chen, Bi, Zhang, Huang, and Hooi}]{liu2025efficient}
Yue Liu, Jiaying Wu, Yufei He, Hongcheng Gao, Hongyu Chen, Baolong Bi, Jiaheng Zhang, Zhiqi Huang, and Bryan Hooi. 2025{\natexlab{c}}.
\newblock Efficient inference for large reasoning models: A survey.
\newblock \emph{arXiv preprint arXiv:2503.23077}.

\bibitem[{Liu et~al.(2025{\natexlab{d}})Liu, Zhai, Du, Chen, Cao, Gao, Wang, Li, Wang, Fang, Zhang, and Hooi}]{guardreasoner-vl}
Yue Liu, Shengfang Zhai, Mingzhe Du, Yulin Chen, Tri Cao, Hongcheng Gao, Cheng Wang, Xinfeng Li, Kun Wang, Junfeng Fang, Jiaheng Zhang, and Bryan Hooi. 2025{\natexlab{d}}.
\newblock \href {http://arxiv.org/abs/2505.11049} {Guardreasoner-vl: Safeguarding vlms via reinforced reasoning}.

\bibitem[{Liu et~al.(2024)Liu, Nie, Tan, Yue, Cui, Wang, Zhu, and Zheng}]{liu2024safety}
Zhendong Liu, Yuanbi Nie, Yingshui Tan, Xiangyu Yue, Qiushi Cui, Chongjun Wang, Xiaoyong Zhu, and Bo~Zheng. 2024.
\newblock Safety alignment for vision language models.
\newblock \emph{arXiv preprint arXiv:2405.13581}.

\bibitem[{Lu et~al.(2023)Lu, Brahman, West, Jang, Chandu, Ravichander, Qin, Ammanabrolu, Jiang, Ramnath et~al.}]{lu2023inference}
Ximing Lu, Faeze Brahman, Peter West, Jaehun Jang, Khyathi Chandu, Abhilasha Ravichander, Lianhui Qin, Prithviraj Ammanabrolu, Liwei Jiang, Sahana Ramnath, et~al. 2023.
\newblock Inference-time policy adapters (ipa): Tailoring extreme-scale lms without fine-tuning.
\newblock \emph{arXiv preprint arXiv:2305.15065}.

\bibitem[{Luong et~al.(2024)Luong, Zhang, Jie, Sun, Jin, and Li}]{Reft}
Trung~Quoc Luong, Xinbo Zhang, Zhanming Jie, Peng Sun, Xiaoran Jin, and Hang Li. 2024.
\newblock Reft: Reasoning with reinforced fine-tuning.
\newblock \emph{arXiv preprint arXiv:2401.08967}, 3.

\bibitem[{Marjanović et~al.(2025)Marjanović, Patel, Adlakha, Aghajohari, BehnamGhader, Bhatia, Khandelwal, Kraft, Krojer, Lù, Meade, Shin, Kazemnejad, Kamath, Mosbach, Stańczak, and Reddy}]{marjanovic2025deepseek}
Sara~Vera Marjanović, Arkil Patel, Vaibhav Adlakha, Milad Aghajohari, Parishad BehnamGhader, Mehar Bhatia, Aditi Khandelwal, Austin Kraft, Benno Krojer, Xing~Han Lù, Nicholas Meade, Dongchan Shin, Amirhossein Kazemnejad, Gaurav Kamath, Marius Mosbach, Karolina Stańczak, and Siva Reddy. 2025.
\newblock \href {http://arxiv.org/abs/2504.07128} {Deepseek-r1 thoughtology: Let's think about llm reasoning}.

\bibitem[{Meta(2024)}]{grattafiori2024llama3herdmodels}
Meta. 2024.
\newblock \href {http://arxiv.org/abs/2407.21783} {The llama 3 herd of models}.

\bibitem[{Mnih et~al.(2015)Mnih, Kavukcuoglu, Silver, Rusu, Veness, Bellemare, Graves, Riedmiller, Fidjeland, Ostrovski et~al.}]{rl2015}
Volodymyr Mnih, Koray Kavukcuoglu, David Silver, Andrei~A Rusu, Joel Veness, Marc~G Bellemare, Alex Graves, Martin Riedmiller, Andreas~K Fidjeland, Georg Ostrovski, et~al. 2015.
\newblock Human-level control through deep reinforcement learning.
\newblock \emph{nature}, 518(7540):529--533.

\bibitem[{Mou et~al.(2025)Mou, Luo, Zhang, and Ye}]{mou2025saro}
Yutao Mou, Yuxiao Luo, Shikun Zhang, and Wei Ye. 2025.
\newblock Saro: Enhancing llm safety through reasoning-based alignment.
\newblock \emph{arXiv preprint arXiv:2504.09420}.

\bibitem[{OpenAI(2024)}]{o1}
OpenAI. 2024.
\newblock Openai o1 system card.
\newblock \emph{arXiv preprint arXiv:2412.16720}.

\bibitem[{Ouyang et~al.(2022)Ouyang, Wu, Jiang, Almeida, Wainwright, Mishkin, Zhang, Agarwal, Slama, Ray et~al.}]{rlhf}
Long Ouyang, Jeffrey Wu, Xu~Jiang, Diogo Almeida, Carroll Wainwright, Pamela Mishkin, Chong Zhang, Sandhini Agarwal, Katarina Slama, Alex Ray, et~al. 2022.
\newblock Training language models to follow instructions with human feedback.
\newblock \emph{Advances in neural information processing systems}, 35.

\bibitem[{Qi et~al.(2021)Qi, Li, Chen, Zhang, Liu, Wang, and Sun}]{qi2021hiddenkillerinvisibletextual}
Fanchao Qi, Mukai Li, Yangyi Chen, Zhengyan Zhang, Zhiyuan Liu, Yasheng Wang, and Maosong Sun. 2021.
\newblock \href {http://arxiv.org/abs/2105.12400} {Hidden killer: Invisible textual backdoor attacks with syntactic trigger}.

\bibitem[{Qiu et~al.(2025)Qiu, Li, Sun, Wei, Xu, Lam, and Yuan}]{qiu2025emerging}
Jianing Qiu, Lin Li, Jiankai Sun, Hao Wei, Zhe Xu, Kyle Lam, and Wu~Yuan. 2025.
\newblock Emerging cyber attack risks of medical ai agents.
\newblock \emph{arXiv preprint arXiv:2504.03759}.

\bibitem[{Qu et~al.(2025)Qu, Li, Su, Sun, Yan, Liu, Cui, Liu, Liang, He et~al.}]{qu2025survey}
Xiaoye Qu, Yafu Li, Zhaochen Su, Weigao Sun, Jianhao Yan, Dongrui Liu, Ganqu Cui, Daizong Liu, Shuxian Liang, Junxian He, et~al. 2025.
\newblock A survey of efficient reasoning for large reasoning models: Language, multimodality, and beyond.
\newblock \emph{arXiv preprint arXiv:2503.21614}.

\bibitem[{Qwen et~al.(2025)Qwen, :, Yang, Yang, Zhang, Hui, Zheng, Yu, Li, Liu, Huang, Wei, Lin, Yang, Tu, Zhang, Yang, Yang, Zhou, Lin, Dang, Lu, Bao, Yang, Yu, Li, Xue, Zhang, Zhu, Men, Lin, Li, Tang, Xia, Ren, Ren, Fan, Su, Zhang, Wan, Liu, Cui, Zhang, and Qiu}]{qwen2025qwen25technicalreport}
Qwen, :, An~Yang, Baosong Yang, Beichen Zhang, Binyuan Hui, Bo~Zheng, Bowen Yu, Chengyuan Li, Dayiheng Liu, Fei Huang, Haoran Wei, Huan Lin, Jian Yang, Jianhong Tu, Jianwei Zhang, Jianxin Yang, Jiaxi Yang, Jingren Zhou, Junyang Lin, Kai Dang, Keming Lu, Keqin Bao, Kexin Yang, Le~Yu, Mei Li, Mingfeng Xue, Pei Zhang, Qin Zhu, Rui Men, Runji Lin, Tianhao Li, Tianyi Tang, Tingyu Xia, Xingzhang Ren, Xuancheng Ren, Yang Fan, Yang Su, Yichang Zhang, Yu~Wan, Yuqiong Liu, Zeyu Cui, Zhenru Zhang, and Zihan Qiu. 2025.
\newblock \href {http://arxiv.org/abs/2412.15115} {Qwen2.5 technical report}.

\bibitem[{Rafailov et~al.(2024)Rafailov, Sharma, Mitchell, Manning, Ermon, and Finn}]{DPO}
Rafael Rafailov, Archit Sharma, Eric Mitchell, Christopher~D Manning, Stefano Ermon, and Chelsea Finn. 2024.
\newblock Direct preference optimization: Your language model is secretly a reward model.
\newblock \emph{Advances in Neural Information Processing Systems}, 36.

\bibitem[{Rein et~al.(2024)Rein, Hou, Stickland, Petty, Pang, Dirani, Michael, and Bowman}]{rein2024gpqa}
David Rein, Betty~Li Hou, Asa~Cooper Stickland, Jackson Petty, Richard~Yuanzhe Pang, Julien Dirani, Julian Michael, and Samuel~R Bowman. 2024.
\newblock Gpqa: A graduate-level google-proof qa benchmark.
\newblock In \emph{First Conference on Language Modeling}.

\bibitem[{Ren et~al.(2024)Ren, Li, Liu, Xie, Lu, Qiao, Sha, Yan, Ma, and Shao}]{ren2024derail}
Qibing Ren, Hao Li, Dongrui Liu, Zhanxu Xie, Xiaoya Lu, Yu~Qiao, Lei Sha, Junchi Yan, Lizhuang Ma, and Jing Shao. 2024.
\newblock Derail yourself: Multi-turn llm jailbreak attack through self-discovered clues.
\newblock \emph{arXiv preprint arXiv:2410.10700}.

\bibitem[{Romero-Arjona et~al.(2025)Romero-Arjona, Valle, Alonso, S{\'a}nchez, Ugarte, Cazalilla, Cambr{\'o}n, Parejo, Arrieta, and Segura}]{romero2025red}
Miguel Romero-Arjona, Pablo Valle, Juan~C Alonso, Ana~B S{\'a}nchez, Miriam Ugarte, Antonia Cazalilla, Vicente Cambr{\'o}n, Jos{\'e}~A Parejo, Aitor Arrieta, and Sergio Segura. 2025.
\newblock Red teaming contemporary ai models: Insights from spanish and basque perspectives.
\newblock \emph{arXiv preprint arXiv:2503.10192}.

\bibitem[{Russinovich et~al.(2024)Russinovich, Salem, and Eldan}]{russinovich2024crescendo}
Mark Russinovich, Ahmed Salem, and Ronen Eldan. 2024.
\newblock Great, now write an article about that: The crescendo multi-turn llm jailbreak attack.
\newblock \emph{arXiv preprint arXiv:2404.01833}.

\bibitem[{Shah et~al.(2023)Shah, Pour, Tagade, Casper, Rando et~al.}]{shah2023scalable}
Rusheb Shah, Soroush Pour, Arush Tagade, Stephen Casper, Javier Rando, et~al. 2023.
\newblock Scalable and transferable black-box jailbreaks for language models via persona modulation.
\newblock \emph{arXiv preprint arXiv:2311.03348}.

\bibitem[{Shen et~al.(2023)Shen, Jin, Huang, Liu, Dong, Guo, Wu, Liu, and Xiong}]{shen2023large}
Tianhao Shen, Renren Jin, Yufei Huang, Chuang Liu, Weilong Dong, Zishan Guo, Xinwei Wu, Yan Liu, and Deyi Xiong. 2023.
\newblock Large language model alignment: A survey.
\newblock \emph{arXiv preprint arXiv:2309.15025}.

\bibitem[{Shi et~al.(2024)Shi, Shen, Huang, Li, Leng, Jin, Liu, Wu, Guo, Yu, Shi, Jiang, and Xiong}]{shi2024largelanguagemodelsafety}
Dan Shi, Tianhao Shen, Yufei Huang, Zhigen Li, Yongqi Leng, Renren Jin, Chuang Liu, Xinwei Wu, Zishan Guo, Linhao Yu, Ling Shi, Bojian Jiang, and Deyi Xiong. 2024.
\newblock \href {http://arxiv.org/abs/2412.17686} {Large language model safety: A holistic survey}.

\bibitem[{Shumailov et~al.(2021)Shumailov, Zhao, Bates, Papernot, Mullins, and Anderson}]{9581273}
Ilia Shumailov, Yiren Zhao, Daniel Bates, Nicolas Papernot, Robert Mullins, and Ross Anderson. 2021.
\newblock \href {https://doi.org/10.1109/EuroSP51992.2021.00024} {Sponge examples: Energy-latency attacks on neural networks}.
\newblock In \emph{2021 IEEE European Symposium on Security and Privacy}, pages 212--231.

\bibitem[{Sun et~al.(2024)Sun, Zhang, Yang, Zou, and Li}]{sun2024multi}
Xiongtao Sun, Deyue Zhang, Dongdong Yang, Quanchen Zou, and Hui Li. 2024.
\newblock Multi-turn context jailbreak attack on large language models from first principles.
\newblock \emph{arXiv preprint arXiv:2408.04686}.

\bibitem[{Sutton et~al.(1998)Sutton, Barto et~al.}]{rl1998}
Richard~S Sutton, Andrew~G Barto, et~al. 1998.
\newblock \emph{Reinforcement learning: An introduction}, volume~1.
\newblock MIT press Cambridge.

\bibitem[{Team et~al.(2025)Team, Du, Gao, Xing, Jiang, Chen, Li, Xiao, Du, Liao et~al.}]{kimi1.5}
Kimi Team, Angang Du, Bofei Gao, Bowei Xing, Changjiu Jiang, Cheng Chen, Cheng Li, Chenjun Xiao, Chenzhuang Du, Chonghua Liao, et~al. 2025.
\newblock Kimi k1. 5: Scaling reinforcement learning with llms.
\newblock \emph{arXiv preprint arXiv:2501.12599}.

\bibitem[{Team(2024{\natexlab{a}})}]{qvq}
Qwen Team. 2024{\natexlab{a}}.
\newblock Qvq: To see the world with wisdom.
\newblock \emph{https://qwenlm.github.io/blog/qvq-72b-preview/}.

\bibitem[{Team(2024{\natexlab{b}})}]{qwq}
Qwen Team. 2024{\natexlab{b}}.
\newblock Qwq: Reflect deeply on the boundaries of the unknown.
\newblock \emph{https://qwenlm.github.io/blog/qwq-32b-preview/}.

\bibitem[{Upadhayay et~al.(2025)Upadhayay, Behzadan et~al.}]{upadhayay2025x}
Bibek Upadhayay, Vahid Behzadan, et~al. 2025.
\newblock X-guard: Multilingual guard agent for content moderation.
\newblock \emph{arXiv preprint arXiv:2504.08848}.

\bibitem[{Wang et~al.(2022{\natexlab{a}})Wang, Ping, Xiao, Xu, Patwary, Shoeybi, Li, Anandkumar, and Catanzaro}]{wang2022exploring}
Boxin Wang, Wei Ping, Chaowei Xiao, Peng Xu, Mostofa Patwary, Mohammad Shoeybi, Bo~Li, Anima Anandkumar, and Bryan Catanzaro. 2022{\natexlab{a}}.
\newblock Exploring the limits of domain-adaptive training for detoxifying large-scale language models.
\newblock \emph{Advances in Neural Information Processing Systems}, 35:35811--35824.

\bibitem[{Wang et~al.(2025{\natexlab{a}})Wang, Qin, Shen, Wang, Cheng, and Tao}]{wang2025leveraging}
Haoyu Wang, Zeyu Qin, Li~Shen, Xueqian Wang, Minhao Cheng, and Dacheng Tao. 2025{\natexlab{a}}.
\newblock Leveraging reasoning with guidelines to elicit and utilize knowledge for enhancing safety alignment.
\newblock \emph{arXiv preprint arXiv:2502.04040}.

\bibitem[{Wang et~al.(2023)Wang, Xu, Lan, Hu, Lan, Lee, and Lim}]{wang2023plan}
Lei Wang, Wanyu Xu, Yihuai Lan, Zhiqiang Hu, Yunshi Lan, Roy Ka-Wei Lee, and Ee-Peng Lim. 2023.
\newblock Plan-and-solve prompting: Improving zero-shot chain-of-thought reasoning by large language models.
\newblock \emph{arXiv preprint arXiv:2305.04091}.

\bibitem[{Wang et~al.(2024{\natexlab{a}})Wang, Zhang, Li, Tan, Wang, Ren, Jiang, and Qiu}]{wang2024inferaligner}
Pengyu Wang, Dong Zhang, Linyang Li, Chenkun Tan, Xinghao Wang, Ke~Ren, Botian Jiang, and Xipeng Qiu. 2024{\natexlab{a}}.
\newblock Inferaligner: Inference-time alignment for harmlessness through cross-model guidance.
\newblock \emph{arXiv preprint arXiv:2401.11206}.

\bibitem[{Wang et~al.(2022{\natexlab{b}})Wang, Kordi, Mishra, Liu, Smith, Khashabi, and Hajishirzi}]{wang2022self}
Yizhong Wang, Yeganeh Kordi, Swaroop Mishra, Alisa Liu, Noah~A Smith, Daniel Khashabi, and Hannaneh Hajishirzi. 2022{\natexlab{b}}.
\newblock Self-instruct: Aligning language models with self-generated instructions.
\newblock \emph{arXiv preprint arXiv:2212.10560}.

\bibitem[{Wang et~al.(2022{\natexlab{c}})Wang, Mishra, Alipoormolabashi, Kordi, Mirzaei, Arunkumar, Ashok, Dhanasekaran, Naik, Stap et~al.}]{wang2022super}
Yizhong Wang, Swaroop Mishra, Pegah Alipoormolabashi, Yeganeh Kordi, Amirreza Mirzaei, Anjana Arunkumar, Arjun Ashok, Arut~Selvan Dhanasekaran, Atharva Naik, David Stap, et~al. 2022{\natexlab{c}}.
\newblock Super-naturalinstructions: Generalization via declarative instructions on 1600+ nlp tasks.
\newblock \emph{arXiv preprint arXiv:2204.07705}.

\bibitem[{Wang et~al.(2025{\natexlab{b}})Wang, Tu, Wang, Wu, Mei, Bartoldson, Kailkhura, and Xie}]{wang2025star}
Zijun Wang, Haoqin Tu, Yuhan Wang, Juncheng Wu, Jieru Mei, Brian~R Bartoldson, Bhavya Kailkhura, and Cihang Xie. 2025{\natexlab{b}}.
\newblock Star-1: Safer alignment of reasoning llms with 1k data.
\newblock \emph{arXiv preprint arXiv:2504.01903}.

\bibitem[{Wang et~al.(2024{\natexlab{b}})Wang, Zhang, Li, Eisenschlos, Perot, Wang, Miculicich, Fujii, Shang, Lee, and Pfister}]{wang2024chainoftableevolvingtablesreasoning}
Zilong Wang, Hao Zhang, Chun-Liang Li, Julian~Martin Eisenschlos, Vincent Perot, Zifeng Wang, Lesly Miculicich, Yasuhisa Fujii, Jingbo Shang, Chen-Yu Lee, and Tomas Pfister. 2024{\natexlab{b}}.
\newblock \href {http://arxiv.org/abs/2401.04398} {Chain-of-table: Evolving tables in the reasoning chain for table understanding}.

\bibitem[{Watkins and Dayan(1992)}]{rl1992}
Christopher~JCH Watkins and Peter Dayan. 1992.
\newblock Q-learning.
\newblock \emph{Machine learning}, 8:279--292.

\bibitem[{Wei et~al.(2022)Wei, Wang, Schuurmans, Bosma, Xia, Chi, Le, Zhou et~al.}]{wei2022chain}
Jason Wei, Xuezhi Wang, Dale Schuurmans, Maarten Bosma, Fei Xia, Ed~Chi, Quoc~V Le, Denny Zhou, et~al. 2022.
\newblock Chain-of-thought prompting elicits reasoning in large language models.
\newblock \emph{Advances in neural information processing systems}, 35:24824--24837.

\bibitem[{Welbl et~al.(2021)Welbl, Glaese, Uesato, Dathathri, Mellor, Hendricks, Anderson, Kohli, Coppin, and Huang}]{welbl2021challenges}
Johannes Welbl, Amelia Glaese, Jonathan Uesato, Sumanth Dathathri, John Mellor, Lisa~Anne Hendricks, Kirsty Anderson, Pushmeet Kohli, Ben Coppin, and Po-Sen Huang. 2021.
\newblock Challenges in detoxifying language models.
\newblock \emph{arXiv preprint arXiv:2109.07445}.

\bibitem[{Wen et~al.(2025)Wen, Zhou, Mo, and Chen}]{ThinkGuard}
Xiaofei Wen, Wenxuan Zhou, Wenjie~Jacky Mo, and Muhao Chen. 2025.
\newblock Thinkguard: Deliberative slow thinking leads to cautious guardrails.
\newblock \emph{arXiv preprint arXiv:2502.13458}.

\bibitem[{Weng et~al.(2025)Weng, Lou, Feng, Huang, and Wang}]{ADPO}
Fenghua Weng, Jian Lou, Jun Feng, Minlie Huang, and Wenjie Wang. 2025.
\newblock Adversary-aware dpo: Enhancing safety alignment in vision language models via adversarial training.
\newblock \emph{arXiv preprint arXiv:2502.11455}.

\bibitem[{Wu et~al.(2021)Wu, Ouyang, Ziegler, Stiennon, Lowe, Leike, and Christiano}]{sft}
Jeff Wu, Long Ouyang, Daniel~M Ziegler, Nisan Stiennon, Ryan Lowe, Jan Leike, and Paul Christiano. 2021.
\newblock Recursively summarizing books with human feedback.
\newblock \emph{arXiv preprint arXiv:2109.10862}.

\bibitem[{Wu et~al.(2025{\natexlab{a}})Wu, Xiang, Wang, and Mittal}]{wu2025effectivelycontrollingreasoningmodels}
Tong Wu, Chong Xiang, Jiachen~T. Wang, and Prateek Mittal. 2025{\natexlab{a}}.
\newblock \href {http://arxiv.org/abs/2503.24370} {Effectively controlling reasoning models through thinking intervention}.

\bibitem[{Wu et~al.(2025{\natexlab{b}})Wu, Wang, Du, Jegelka, and Wang}]{wu2025more}
Yuyang Wu, Yifei Wang, Tianqi Du, Stefanie Jegelka, and Yisen Wang. 2025{\natexlab{b}}.
\newblock When more is less: Understanding chain-of-thought length in llms.
\newblock \emph{arXiv preprint arXiv:2502.07266}.

\bibitem[{Xiang et~al.(2024)Xiang, Jiang, Xiong, Ramasubramanian, Poovendran, and Li}]{xiang2024badchain}
Zhen Xiang, Fengqing Jiang, Zidi Xiong, Bhaskar Ramasubramanian, Radha Poovendran, and Bo~Li. 2024.
\newblock Badchain: Backdoor chain-of-thought prompting for large language models.
\newblock \emph{arXiv preprint arXiv:2401.12242}.

\bibitem[{Xu et~al.(2023)Xu, Ma, Wang, Xiao, and Chen}]{xu2023instructions}
Jiashu Xu, Mingyu~Derek Ma, Fei Wang, Chaowei Xiao, and Muhao Chen. 2023.
\newblock Instructions as backdoors: Backdoor vulnerabilities of instruction tuning for large language models.
\newblock \emph{arXiv preprint arXiv:2305.14710}.

\bibitem[{Xu et~al.(2025)Xu, Li, Chen, and Xu}]{xu2025nuclear}
Rongwu Xu, Xiaojian Li, Shuo Chen, and Wei Xu. 2025.
\newblock Nuclear deployed: Analyzing catastrophic risks in decision-making of autonomous llm agents.
\newblock \emph{arXiv preprint arXiv:2502.11355}.

\bibitem[{Yang et~al.(2025)Yang, He, Pan, Jiang, Deng, Yang, Lu, Yin, Rao, Zhu, Zhang, and Chen}]{yang2025r1onevisionadvancinggeneralizedmultimodal}
Yi~Yang, Xiaoxuan He, Hongkun Pan, Xiyan Jiang, Yan Deng, Xingtao Yang, Haoyu Lu, Dacheng Yin, Fengyun Rao, Minfeng Zhu, Bo~Zhang, and Wei Chen. 2025.
\newblock \href {http://arxiv.org/abs/2503.10615} {R1-onevision: Advancing generalized multimodal reasoning through cross-modal formalization}.

\bibitem[{Yao et~al.(2024{\natexlab{a}})Yao, Lou, and Qin}]{10446267}
Hongwei Yao, Jian Lou, and Zhan Qin. 2024{\natexlab{a}}.
\newblock \href {https://doi.org/10.1109/ICASSP48485.2024.10446267} {Poisonprompt: Backdoor attack on prompt-based large language models}.
\newblock In \emph{ICASSP 2024 - 2024 IEEE International Conference on Acoustics, Speech and Signal Processing (ICASSP)}, pages 7745--7749.

\bibitem[{Yao et~al.(2024{\natexlab{b}})Yao, Huang, Wu, Zhang, Wang, Liu, Wang, Song, Feng, Shen, and Tao}]{yao2024mulberryempoweringmllmo1like}
Huanjin Yao, Jiaxing Huang, Wenhao Wu, Jingyi Zhang, Yibo Wang, Shunyu Liu, Yingjie Wang, Yuxin Song, Haocheng Feng, Li~Shen, and Dacheng Tao. 2024{\natexlab{b}}.
\newblock \href {http://arxiv.org/abs/2412.18319} {Mulberry: Empowering mllm with o1-like reasoning and reflection via collective monte carlo tree search}.

\bibitem[{Yao et~al.(2023)Yao, Yu, Zhao, Shafran, Griffiths, Cao, and Narasimhan}]{yao2023treethoughtsdeliberateproblem}
Shunyu Yao, Dian Yu, Jeffrey Zhao, Izhak Shafran, Thomas~L. Griffiths, Yuan Cao, and Karthik Narasimhan. 2023.
\newblock \href {http://arxiv.org/abs/2305.10601} {Tree of thoughts: Deliberate problem solving with large language models}.

\bibitem[{Yao et~al.(2025)Yao, Tong, Wang, Wang, Li, Liu, Teng, and Wang}]{yao2025mousetrap}
Yang Yao, Xuan Tong, Ruofan Wang, Yixu Wang, Lujundong Li, Liang Liu, Yan Teng, and Yingchun Wang. 2025.
\newblock A mousetrap: Fooling large reasoning models for jailbreak with chain of iterative chaos.
\newblock \emph{arXiv preprint arXiv:2502.15806}.

\bibitem[{Ye et~al.(2025)Ye, Rong, Huang, Du, Yu, and Tao}]{vlm_safety_survey}
Mang Ye, Xuankun Rong, Wenke Huang, Bo~Du, Nenghai Yu, and Dacheng Tao. 2025.
\newblock A survey of safety on large vision-language models: Attacks, defenses and evaluations.
\newblock \emph{arXiv preprint arXiv:2502.14881}.

\bibitem[{Yi et~al.(2024)Yi, Liu, Sun, Cong, He, Song, Xu, and Li}]{yi2024jailbreakattacksdefenseslarge}
Sibo Yi, Yule Liu, Zhen Sun, Tianshuo Cong, Xinlei He, Jiaxing Song, Ke~Xu, and Qi~Li. 2024.
\newblock \href {http://arxiv.org/abs/2407.04295} {Jailbreak attacks and defenses against large language models: A survey}.

\bibitem[{Ying et~al.(2025{\natexlab{a}})Ying, Zhang, Jing, Xiao, Zou, Liu, Liang, Zhang, Liu, and Tao}]{ying2025reasoning}
Zonghao Ying, Deyue Zhang, Zonglei Jing, Yisong Xiao, Quanchen Zou, Aishan Liu, Siyuan Liang, Xiangzheng Zhang, Xianglong Liu, and Dacheng Tao. 2025{\natexlab{a}}.
\newblock Reasoning-augmented conversation for multi-turn jailbreak attacks on large language models.
\newblock \emph{arXiv preprint arXiv:2502.11054}.

\bibitem[{Ying et~al.(2025{\natexlab{b}})Ying, Zheng, Huang, Zhang, Zhang, Zou, Liu, Liu, and Tao}]{ying2025towards}
Zonghao Ying, Guangyi Zheng, Yongxin Huang, Deyue Zhang, Wenxin Zhang, Quanchen Zou, Aishan Liu, Xianglong Liu, and Dacheng Tao. 2025{\natexlab{b}}.
\newblock Towards understanding the safety boundaries of deepseek models: Evaluation and findings.
\newblock \emph{arXiv preprint arXiv:2503.15092}.

\bibitem[{Zaremba et~al.(2025)Zaremba, Nitishinskaya, Barak, Lin, Toyer, Yu, Dias, Wallace, Xiao, Heidecke, and Glaese}]{zaremba2025tradinginferencetimecomputeadversarial}
Wojciech Zaremba, Evgenia Nitishinskaya, Boaz Barak, Stephanie Lin, Sam Toyer, Yaodong Yu, Rachel Dias, Eric Wallace, Kai Xiao, Johannes Heidecke, and Amelia Glaese. 2025.
\newblock \href {http://arxiv.org/abs/2501.18841} {Trading inference-time compute for adversarial robustness}.

\bibitem[{Zeng et~al.(2024{\natexlab{a}})Zeng, Liu, Mullins, Peran, Fernandez, Harkous, Narasimhan, Proud, Kumar, Radharapu et~al.}]{Shieldgemma}
Wenjun Zeng, Yuchi Liu, Ryan Mullins, Ludovic Peran, Joe Fernandez, Hamza Harkous, Karthik Narasimhan, Drew Proud, Piyush Kumar, Bhaktipriya Radharapu, et~al. 2024{\natexlab{a}}.
\newblock Shieldgemma: Generative ai content moderation based on gemma.
\newblock \emph{arXiv preprint arXiv:2407.21772}.

\bibitem[{Zeng et~al.(2024{\natexlab{b}})Zeng, Lin, Zhang, Yang, Jia, and Shi}]{zeng2024johnny}
Yi~Zeng, Hongpeng Lin, Jingwen Zhang, Diyi Yang, Ruoxi Jia, and Weiyan Shi. 2024{\natexlab{b}}.
\newblock How johnny can persuade llms to jailbreak them: Rethinking persuasion to challenge ai safety by humanizing llms.
\newblock In \emph{Proceedings of the 62nd Annual Meeting of the Association for Computational Linguistics (Volume 1: Long Papers)}, pages 14322--14350.

\bibitem[{Zhang et~al.(2025{\natexlab{a}})Zhang, Lei, Liu, Wang, Long, Yang, Zhao, Hua, Ma, Wang et~al.}]{zhang2025safety}
Wenjing Zhang, Xuejiao Lei, Zhaoxiang Liu, Ning Wang, Zhenhong Long, Peijun Yang, Jiaojiao Zhao, Minjie Hua, Chaoyang Ma, Kai Wang, et~al. 2025{\natexlab{a}}.
\newblock Safety evaluation of deepseek models in chinese contexts.
\newblock \emph{arXiv preprint arXiv:2502.11137}.

\bibitem[{Zhang et~al.(2025{\natexlab{b}})Zhang, Zeng, Li, Huang, Deng, and Dong}]{zhang2025realsafe}
Yichi Zhang, Zihao Zeng, Dongbai Li, Yao Huang, Zhijie Deng, and Yinpeng Dong. 2025{\natexlab{b}}.
\newblock Realsafe-r1: Safety-aligned deepseek-r1 without compromising reasoning capability.
\newblock \emph{arXiv preprint arXiv:2504.10081}.

\bibitem[{Zhang et~al.(2025{\natexlab{c}})Zhang, Zhang, Huang, Xia, Fang, Yang, Duan, Yan, Dong, and Zhu}]{zhang2025stair}
Yichi Zhang, Siyuan Zhang, Yao Huang, Zeyu Xia, Zhengwei Fang, Xiao Yang, Ranjie Duan, Dong Yan, Yinpeng Dong, and Jun Zhu. 2025{\natexlab{c}}.
\newblock Stair: Improving safety alignment with introspective reasoning.
\newblock \emph{arXiv preprint arXiv:2502.02384}.

\bibitem[{Zhang et~al.(2024)Zhang, Chen, Zheng, Gao, Zheng, Fu, Yin, Jin, Qiao, Huang et~al.}]{zhang2024spa}
Yongting Zhang, Lu~Chen, Guodong Zheng, Yifeng Gao, Rui Zheng, Jinlan Fu, Zhenfei Yin, Senjie Jin, Yu~Qiao, Xuanjing Huang, et~al. 2024.
\newblock Spa-vl: A comprehensive safety preference alignment dataset for vision language model.
\newblock \emph{arXiv preprint arXiv:2406.12030}.

\bibitem[{Zhao et~al.(2025)Zhao, Wu, Zhang, and Vasilakos}]{zhao2025shadowcot}
Gejian Zhao, Hanzhou Wu, Xinpeng Zhang, and Athanasios~V Vasilakos. 2025.
\newblock Shadowcot: Cognitive hijacking for stealthy reasoning backdoors in llms.
\newblock \emph{arXiv preprint arXiv:2504.05605}.

\bibitem[{Zhao et~al.(2024)Zhao, Jia, Guo, Gan, Xu, Wu, Fu, Feng, Pan, and Tuan}]{zhao2024survey}
Shuai Zhao, Meihuizi Jia, Zhongliang Guo, Leilei Gan, Xiaoyu Xu, Xiaobao Wu, Jie Fu, Yichao Feng, Fengjun Pan, and Luu~Anh Tuan. 2024.
\newblock A survey of backdoor attacks and defenses on large language models: Implications for security measures.
\newblock \emph{Authorea Preprints}.

\bibitem[{Zhou et~al.(2025)Zhou, Liu, Zhao, Jangam, Srinivasa, Liu, Song, and Wang}]{zhou2025hidden}
Kaiwen Zhou, Chengzhi Liu, Xuandong Zhao, Shreedhar Jangam, Jayanth Srinivasa, Gaowen Liu, Dawn Song, and Xin~Eric Wang. 2025.
\newblock The hidden risks of large reasoning models: A safety assessment of r1.
\newblock \emph{arXiv preprint arXiv:2502.12659}.

\bibitem[{Zhou(2020)}]{tocxic_roberta}
Xuhui Zhou. 2020.
\newblock \emph{Challenges in automated debiasing for toxic language detection}.
\newblock University of Washington.

\bibitem[{Zhu et~al.(2025{\natexlab{a}})Zhu, Yan, Wang, Yin, and Sha}]{zhu2025reasoning}
Junda Zhu, Lingyong Yan, Shuaiqiang Wang, Dawei Yin, and Lei Sha. 2025{\natexlab{a}}.
\newblock Reasoning-to-defend: Safety-aware reasoning can defend large language models from jailbreaking.
\newblock \emph{arXiv preprint arXiv:2502.12970}.

\bibitem[{Zhu et~al.(2025{\natexlab{b}})Zhu, Zhang, Zhang, Wang, Wu, Xu, and Wu}]{zhu2025bot}
Zihao Zhu, Hongbao Zhang, Mingda Zhang, Ruotong Wang, Guanzong Wu, Ke~Xu, and Baoyuan Wu. 2025{\natexlab{b}}.
\newblock Bot: Breaking long thought processes of o1-like large language models through backdoor attack.
\newblock \emph{arXiv preprint arXiv:2502.12202}.

\end{thebibliography}

\appendix

\end{document}